\documentclass[letterpaper]{article} 
\usepackage{aaai2026}  
\usepackage{times}  
\usepackage{helvet}  
\usepackage{courier}  
\usepackage[hyphens]{url}  
\usepackage{graphicx} 
\usepackage{natbib}  
\usepackage{caption} 
\usepackage{algorithm}
\usepackage{algorithmic}
\newcommand{\clip}{CLIP}
\newcommand{\mr}[1]{\mathit{#1}}

\newcommand{\loss}{\mathcal{L}}
\newcommand{\mypar}[1]{\vspace{0.0\baselineskip}\noindent\textbf{#1}\,}

\newcommand{\xx}{\mathbf{x}}

\newcommand{\PP}{\mathbf{P}}
\newcommand{\QQ}{\mathbf{Q}}
\newcommand{\SSS}{\mathbf{S}}

\newcommand{\data}{\mathcal{D}}

\newcommand{\tr}[1]{{#1^{\top}}}
\usepackage{makecell}
\usepackage{amsmath}
\usepackage{amssymb}
\usepackage{bm,upgreek}
\usepackage{booktabs}
\usepackage{multirow}
\usepackage{color}
\usepackage{pifont}
\usepackage{newfloat}
\usepackage{listings}
\DeclareCaptionStyle{ruled}{labelfont=normalfont,labelsep=colon,strut=off} 
\floatstyle{ruled}
\newfloat{listing}{tb}{lst}{}
\floatname{listing}{Listing}
\title{Federated CLIP for Resource-Efficient Heterogeneous Medical Image Classification}
\author{Yihang Wu\textsuperscript{\rm 1}, Ahmad Chaddad\textsuperscript{\rm 1,2}\thanks{Correspondence: ahmad8chaddad@gmail.com}
}
\affiliations{
    \textsuperscript{\rm 1}AIPM, School of Artificial Intelligence, Guilin University of Electronic Technology, China\\
    \textsuperscript{\rm 2}The Imaging, Vision and Artificial Intelligence Laboratory, École de Technologie Supérieure, Canada\\
}

\begin{document}

\maketitle

\begin{abstract}
Despite the remarkable performance of deep models in medical imaging, they still require source data for training, which limits their potential in light of privacy concerns. Federated learning (FL), as a decentralized learning framework that trains a shared model with multiple hospitals (a.k.a., FL clients), provides a feasible solution. However, data heterogeneity and resource costs hinder the deployment of FL models, especially when using vision language models (VLM). To address these challenges, we propose a novel contrastive language-image pre-training (CLIP) based FL approach for medical image classification (FedMedCLIP). Specifically, we introduce a masked feature adaptation module (FAM) as a communication module to reduce the communication load while freezing the CLIP encoders to reduce the computational overhead. Furthermore, we propose a masked multi-layer perceptron (MLP) as a private local classifier to adapt to the client tasks. Moreover, we design an adaptive Kullback-Leibler (KL) divergence-based distillation regularization method to enable mutual learning between FAM and MLP. Finally, we incorporate model compression to transmit the FAM parameters while using ensemble predictions for classification. Extensive experiments on four publicly available medical datasets demonstrate that our model provides feasible performance (e.g., 8\% higher compared to second best baseline on ISIC2019) with reasonable resource cost (e.g., 120$\times$ faster than FedAVG).
\end{abstract}

\begin{links}
    \link{Code}{https://github.com/AIPMLab/FedMedCLIP}

\end{links}

\section{Introduction}

With the development of deep learning (DL) in medical imaging, privacy concerns have hindered collaboration between organizations \cite{zeng2024tackling}. Federated learning (FL) has emerged as a solution that allows decentralized training without sacrificing patient privacy, advancing medical artificial intelligence (AI) applications while maintaining data security \cite{zhu2024stealing,chaddad2023federated,chaddad2023explainable}. For example, in \cite{mcmahan2017communication}, they propose a simple federated aggregation method called FedAVG, and the experimental results show that it provides feasible performance without accessing raw data from other clients. Similarly to FedAVG, in \cite{li2020federated}, they propose using a regularization term to measure the discrepancies between the global and local models, thus improving performance in local clients (FedProx).

However, despite advances in FL, it faces two challenges, 1) heterogeneous data between clients, and 2) communication and computational load during local training and global aggregation \cite{wu2024facmic}. Basically, heterogeneous data (e.g., feature shifts) can lead to severe performance degradation on local clients, while communication and computational costs can prohibit the implementation of FL systems in low-resource devices. This is especially important in the era of vision language models such as contrastive language image pre-training (CLIP), which require large computational and communication loads (e.g., $\sim 10^8$ parameters inside) \cite{radford2021learning}. This leads to open question that \textit{``How to adapt the CLIP models effectively in FL context with reasonable cost and feasible generalization performance.''}

Recent studies have introduced parameter efficient pre-training (PEFT) techniques to adapt the CLIP into FL frameworks. For example, in FedCLIP \cite{lu2023fedclip}, they propose an adapter as a communication module while inserting it after the CLIP encoders. Furthermore, they freeze the CLIP encoder parameters to save computational overhead. The experimental results on the multi-domain dataset OfficeHome indicate that it achieves feasible performance compared to FedAVG, while reducing resource costs. Similarly, prompt learning based approaches such as promptFL \cite{guo2023promptfl}, FedAPT \cite{su2024federated} have demonstrated remarkable performance for natural image classification tasks in heterogeneous settings. Despite these advancements, the key idea remains the same: \textit{Balancing utility and model size in FL.} However, none of these studies involved experiments with medical data, where heterogeneity is common in medical imaging (e.g., modality). Furthermore, in \cite{huix2024natural}, they suggest that CLIP exhibits poor recall rate $\sim 50\%$ for the classification of skin cancer, indicating that there exists a large domain gap between natural and medical.

While vanilla CLIP underperforms in medical datasets, its pre-trained backbone remains a valuable starting point. Motivated by previous challenges, our goal is to build a practical adaptation framework that unlocks this potential for medical use without full re-pretraining. Specifically, we use pre-trained CLIP as feature extractors (the encoders are frozen), while introducing a novel masked feature adaptation module (FAM) as communication module. Furthermore, we propose a masked MLP for local tasks, while keeping it local without aggregation. To improve the prediction similarity between FAM and MLP, we incorporate a class-wise KL based distillation approach to minimize the differences between the predicted probabilities of FAM and MLP. To enrich the model predictions, we use ensemble predictions obtained from both FAM and MLP. In addition, we introduce model compression to decrease the communication overhead. Finally, the FAMs are aggregated using a simple average aggregation technique. The contributions of this paper can be summarized as follows.
\begin{enumerate}
    \item \textit{Algorithm:} We propose a novel CLIP-based FL framework for medical image classification tasks (FedMedCLIP). Specifically, we propose a masked FAM, a masked local classifier and a class-wise KL based regularization technique to improve the performance in heterogeneous data setting with reasonable computational and communication cost. Furthermore, we introduce a model compression technique to compress the model parameters before sending them to the global server.
    \item \textit{Empirical analysis:} We perform extensive experiments on four medical datasets (e.g., brain tumor, skin cancer) to show the usefulness of the proposed approach in: (i) robustness to heterogeneity FL, (ii) generalization ability to unseen clients, (iii) efficiency of computation and communication procedure, (iv) adaptability with different network architectures and (v) resilience against adversarial perturbations.
\end{enumerate}

\section{Related Work}

\mypar{Federated learning.} Federated learning serves as a fundamental framework for training robust models without sharing raw data. For example, FedProx introduced a regularization term to measure the discrepancies between the global and local models, thereby improving performance in local clients \cite{li2020federated}. Similar to FedProx, the differences between the local model of the previous round and the global model are calculated using cosine similarities to optimize the local models are proposed in MOON \cite{li2021model}. FedFocal extended the focal loss in FL framework to sovle the class imbalance challenge existed in medical datasets \cite{fedfocal}. In addition, FedProto focused on maximizing
feature-level consistency between local and global models to improve the overall performance \cite{tan2022fedproto}. SCAFFOLD introduced global gradient calibration and controlled variates to correct local optimization directions \cite{karimireddy2020scaffold}. There are also recent studies devoted to FL designs \cite{qin2023fedapen,yang2024fedfed,yu2025modeling}. However, these methods are not suitable for VLMs such as CLIP because they require a large amount of parameters to be transmitted.

\mypar{CLIP.} For example, FACMIC extended FedCLIP to medical classification by introducing domain adaptation technique and a novel adapter as communication module \cite{wu2024facmic}. However, finding a publicly available source domain is a challenge for real-world applications. PromptFL demonstrated the usefulness of sharing prompts instead of models in the FL context \cite{guo2023promptfl}. The experiments show that promptFL yields remarkable performance compared to vanilla methods such as FedAVG. Furthermore, based on promptFL, FedAPT proposed adaptive prompts for local clients (i.e., local prompts with global prompts) for global aggregation \cite{su2024federated}, and the experimental results showed that their method provides comparable performance compared to the centralized approach. However, their method introduces considerable computational overhead during inference. Shi et al. proposed client-side knowledge distillation and server-side federated contrastive learning guided by CLIP to address the heterogeneity existing in local clients \cite{shi2024clip}. However, they do not validate their model under the feature shift condition (e.g., OfficeHome). In addition, FAA-CLIP adapted the CLIP with an adapter and a domain adversarial classifier in FL \cite{10902405}. Experimental results show that it achieves feasible classification performance.

Unlike previous static methods like FACMIC that rely on static aggregation, our approach dynamically learns both shared and client-specific features using a masked FAM, client-specific MLP, and class-wise KL loss to balance consistency. This design jointly enhances personalization and generalization, achieving superior performance.

\section{Methodology}

\begin{figure}
    \centering
    \includegraphics[width=0.99\linewidth]{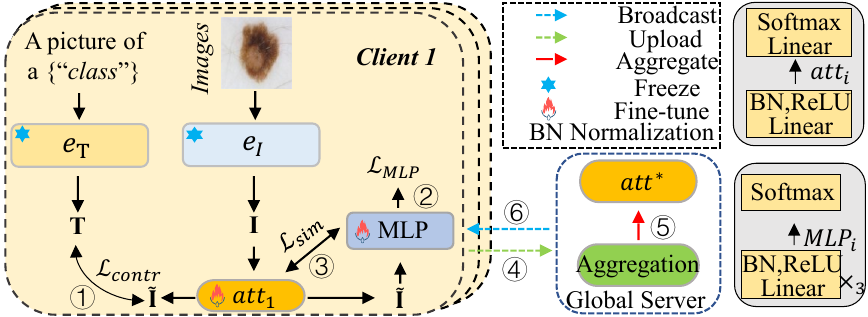}
    \caption{Framework of our approach. The number from \textcircled{\footnotesize{1}} to \textcircled{\footnotesize{6}} indicates the corresponding terms of Eq. (\ref{EQ:contr}), Eq. (\ref{EQ:AMLP}), Eq. (\ref{EQ:sim}), upload, Eq. (\ref{EQ:agg}) and download, respectively.}
    \label{fig:Pipeline}
\end{figure}

\mypar{General framework.} Figure \ref{fig:Pipeline} shows the framework of our method. It consists of three parts: 1) local training and inference, 2) model compression and decompression, and 3) global aggregation. 

\mypar{Local training and inference.} For feature extraction, we use a pretrained \clip{} model, comprising an image encoder $e_I(\cdot)$ and a text encoder $e_T(\cdot)$ for each client $C_i$. For a training example $\xx_j \in \data_{i}^{\mr{train}}$, we denote as $\mathbf{I}_j = e_I(\xx_j) \in \mathrm{R}^D$ the $D$-dimensional vector of image features. For text features, we use the standard prompt ``{\texttt{a picture of a \{class\}}}'' as input to the text encoder to obtain the features $\mathbf{T}_j = e_T(\xx_j) \in \mathrm{R}^D$. To effectively adapt the CLIP model for specific tasks, motivated by \cite{gao2024clip}, we introduce a FAM to the CLIP, denoted $att_i(\cdot)$. This FAM takes as input image features $\mathbf{I}$ and returns an attention mask $att_i(\mathbf{I})\in [0,1]^D$. This mask is then used to generate the features of the masked images $\mathbf{\tilde{I}} = att_i(\mathbf{I}) \otimes \mathbf{I}$, where $\otimes$ is the Hadamard product (element-wise). After obtaining image and text features, the probability that an example $\xx_j$ belongs to a class $c$ can be computed using the cosine similarity $s_{j,c}$ between the image features of $\xx_j$ and the text features $\mathbf{T}_c$ corresponding to the prompt of $c$:
\begin{equation}\footnotesize\footnotesize
\label{eq:CLIP-classif}
p(\mathrm{Y}\!=\!c \, | \, \xx_j) \, =\,  \frac{\exp(s_{j,c}/\tau)}{\sum_{c'=1}^K \exp(s_{j,c'}/\tau)}, \ \ \text{with }
s_{j,c} = \frac{\langle\mathbf{\tilde{I}}_j, \mathbf{T}_c\rangle}{\|\mathbf{\tilde{I}}_j\| \!\cdot\!\|\mathbf{T}_c\|}
\end{equation}
where $\tau$ is the pre-defined softmax temperature parameter. 

To reduce the computational load, the \clip{} encoders are frozen, and we introduce a masked linear layer to perform transformation. The motivation of using masks is that a masked FAM helps to efficiently learn sparse but dominant feature representations across clients \cite{kim2024spafl}. Suppose $W$ and $b$ are the weight matrix and bias, respectively. First, we calculate the mean magnitude $u_i$ of each layer:
\begin{equation}\footnotesize
    u_i = \frac{1}{n_{in}}\sum_{j=1}^{n_{in}}|W_{i,j}|
\end{equation}

After obtaining $u_i$, we generate the mask as follows:
\begin{equation}\footnotesize
    {m_i}=\mathcal{S}(u_i-\kappa_i
)= \left\{ {\begin{array}{*{20}{c}}
{1,\, \, \text{if} \, \, \, \, {\rm{  }}{{\rm{u}}_i} \ge {\kappa _i}}\\
{0, \, \, \text{if} \, \, \, \, {\rm{  }}{{\rm{u}}_i} < {\kappa _i}}
\end{array}} \right.
\end{equation}
where $\kappa$ is a learnable threshold, $\mathcal{S}$ is a sign function. Finally, the $W$ are masked using the masks $\mathcal{M}$ and each linear layer can be formulated as follows:
\begin{equation}\footnotesize\label{eq:MaskMLP}
    \hat{W}=W\odot (\mathcal{M}\cdot\mathbf{1}^{\text{T}})\,,\, y=\hat{W}x \,+\, (b\odot\mathcal{M})
\end{equation}
where $x$ indicates the input.
 
Specifically, the local FAMs are trained by minimizing a contrastive loss $\loss_{\mr{contr}}$ that pushes the image and text features from the same training example together and pulls apart the non-matching ones. In practice, $\loss_{\mr{contr}}$ is calculated on batches of size $B$. Following \cite{wu2024facmic}, let $\SSS$ be the $B\!\times\!B$ matrix where $s_{j,j'}$ is the cosine similarity between the image features $\mathbf{\tilde{I}}_j$ and $\mathbf{T}_{j'}$ as measured in Eq (\ref{eq:CLIP-classif}). We compute an image probability matrix $\PP = \mr{Softmax}(\SSS/\tau) \in [0,1]^{B \times B}$ and a text probability matrix $\QQ = \mr{Softmax}(\tr{\SSS}/\tau) \in [0,1]^{B \times B}$. The contrastive loss is then formulated as follows:
\begin{equation}\footnotesize
\label{EQ:contr}
\loss_{\mr{contr}} \, = \, -\frac{1}{B}\sum_{j=1}^B \frac{1}{2}\Big(\log p_{j,j} \, + \, \log q_{j,j}\Big). 
\end{equation}

Although a global model can provide robust performance, global aggregation can reduce the client-specific features learned by FAM, which can degrade the performance in clients. Therefore, we propose a MLP as the local classifier for all clients. Unlike traditional MLPs, the proposed MLP replaces the linear layer with a masked linear layer, as defined in Eq. \ref{eq:MaskMLP}. This allows the MLP to focus on task-specific features while ensuring the sparsity of the model structure. Specifically, the local MLP generates raw logits $\mathbf{O}^m\in\mathbb{R}^{B\times C}$, then it goes through a Softmax function to output the probability matrix $\textbf{P}^{m} \in [0,1]^{B\times C}$ according to the number of classes $C$ using $\mathbf{\tilde{I}}$. Then, we measure the $\loss_{MLP}$ as follows:
\begin{equation}\footnotesize\label{EQ:AMLP}
\loss_{MLP} = - \frac{1}{B}\sum\limits_{j = 1}^B {\mathcal{L}_{CE}({{{p}}_j},} y) \end{equation}

If the quality of the clients data is low, it can degrade the performance of the local MLP. Therefore, we use the KL divergence to let the global FAM and the local MLP learn mutually. Specifically, we designed a class-wise KL regularization for local training because applying KL without considering class information can lead to misadaptation due to client heterogeneity. This helps the FAM and MLP learn class-wise information mutually (i.e., enrich the FAM with client-specific features and vice versa) \cite{cui2024decoupled}. Let $\textbf{P}^{v} \in [0,1]^{B\times C}$ be the predicted similarities of the CLIP decoder, the loss of consistency regulation ($\loss_{sim}$) can be formulated as follows.
\begin{equation}\footnotesize\footnotesize\label{EQ:sim}
    \mathcal{L}_{sim} = \frac{1}{{2C}}\left( {\sum\limits_{c = 1}^C {\sum\limits_{i = 1}^B {\hat q_i^{(c)}} } \log \frac{{\hat q_i^{(c)}}}{{\hat p_i^{(c)}}} + \sum\limits_{c = 1}^C {\sum\limits_{i = 1}^B {\hat p_i^{(c)}} } \log \frac{{\hat p_i^{(c)}}}{{\hat q_i^{(c)}}}} \right)
\end{equation}
where $\hat{q}_i^{(c)}$ and $\hat{p}_i^{(c)}$ are defined as:
\begin{equation}\footnotesize\footnotesize\label{EQ:T}
    \hat q_i^{(c)} = \frac{{\exp (s_i^{(c)}/T)}}{{\sum\nolimits_{b = 1}^B {\exp (s_b^{(c)}/T)} }},\hat p_i^{(c)} = \frac{{\exp (o_i^{(c)}/T)}}{{\sum\nolimits_{b = 1}^B {\exp (o_b^{(c)}/T)} }}
\end{equation}
where $T$ is a scale parameter. $s$ and $o$ are the logits from FAM and MLP. To more precisely regularize FAM and MLP, we design a dynamic weight parameter $\varpi$ to balance the impact of Eq. \ref{EQ:sim} as follows:
\begin{equation}\footnotesize\footnotesize
\mathcal{L}_{sim} = \frac{1}{{C}}\left( {\sum\limits_{c = 1}^C {\sum\limits_{i = 1}^B {( {\varpi \hat q_i^{(c)}\log \frac{{\hat q_i^{(c)}}}{{\hat p_i^{(c)}}} + \left( {1 - \varpi } \right)\hat p_i^{(c)}\log \frac{{\hat p_i^{(c)}}}{{\hat q_i^{(c)}}}} )} } } \right)  
\end{equation}

Here, we propose a training-agnositic $\varpi$ as follows.
\begin{equation}\footnotesize
   \varpi  = \frac{{\mathcal{H}\left( {{p^v}} \right)}}{{\mathcal{H}\left( {{p^m}} \right) + \mathcal{H}\left( {{p^v}} \right)}}
\end{equation}
where $\mathcal{H}$ is the entropy function. By minimizing $\mathcal{L}_{sim}$, it can help the private model to learn from the global model, and vice versa. This enhances their overall performance and consistency.

Finally, each local model optimizes the following loss function with a hyperparameter $\lambda$:
\begin{equation}\footnotesize\label{EQ:8}
    \loss = \loss_{contr} + \loss_{MLP} + \lambda\cdot \loss_{sim}
\end{equation}
 
During inference, since the global model benefits more from knowledgeable clients while the local classifier fits more about local tasks, we propose to use ensemble predictions
to improve the classification ability. Basically, for each sample, the final predicted probability $p^{ens}$ can be measured as:
\begin{equation}\footnotesize
     p^{ens} = \varpi \cdot p^{MLP} + (1-\varpi) \cdot p^{FAM}
\end{equation}

\mypar{Model compression and decompression.} We compress the model parameters before sending it to the global server. Specifically, the compression pipeline converts model parameters (i.e., weights) to float16, packs metadata using network byte order, serializes parameters as binary data, and applies zlib compression \cite{gailly2004zlib} to ensure storage efficiency. When local clients receive the compressed model parameters, decompression is performed to update the local model. Decompression reverses the compression steps by decompressing the zlib data, parsing metadata in big-endian format, converting binary data to tensors, adjusting precision from float16 to float32, and restoring parameters to the model state dictionary.

\begin{table}[!ht] 
\footnotesize
    \setlength{\tabcolsep}{1.89pt}
    \renewcommand{\arraystretch}{0.65}
    
    \begin{tabular}{c|cccccc|c|c}
    \toprule 
     Method&\multicolumn{6}{|c|}{Client}&Global&\multirow{2}{*}{AVG}  \\
     \cmidrule(l{3pt}r{3pt}){1-1}\cmidrule(l{3pt}r{3pt}){2-8}
    AS&$C_1$ & $C_2$& $C_3$& $C_4$&$C_5$&$C_6$& $C_{glo}$& \\
    \midrule
    CLIP$_{zs}$&31.95&23.98&24.31&17.71&20.12&33.47&17.17&24.1 \\
Individual&74.82&52.03&74.42&\underline{67.71}&69.14&59.48&17.17&59.25 \\
FedAVG&\underline{78.35}&\underline{60.47}&{76.39}&58.33&75.00&{71.77}&\underline{84.54}&{72.12} \\
LoRA&72.10&59.45&\underline{76.50}&55.21&71.29&\underline{72.18}&82.26&69.86 \\
PromptFL&74.21&48.26&\underline{76.50}&62.50&76.95&69.76&84.42&70.37 \\
CocoOp&73.94&49.13&74.77&61.46&75.59&68.35&82.63&69.41 \\
FedCLIP&69.54&56.98&74.88&57.29&70.12&71.37&79.91&68.58\\
FACMIC&69.19&56.10&74.54&60.42&67.58&66.94&74.23&67.0 \\
LP++&67.87&27.18&68.17&59.38&74.22&57.66&77.52&61.71 \\
FedAPT&77.15 & 51.09 & 73.80 & {67.65} & \textbf{81.70} & 68.60 & \textbf{85.43}& \underline{72.21}\\
Ours&\textbf{84.45}&\textbf{71.92}&\textbf{82.69}&\textbf{84.31}&\underline{79.00}&\textbf{79.40}&81.01&\textbf{80.4}\\
    \bottomrule
    \end{tabular}
\caption{Accuracy (\%) in ISIC2019. \textbf{Bold} means the best, while \underline{Underline} indicates the second best. AVG is the average value of the client accuracy and global accuracy.}
    \label{tab:ISIC2019}
\end{table}

\mypar{Global aggregation.} We use a simple average aggregation to provide an unbiased aggregated global model. Specifically, the local clients compress their local models and then send them to the global server, while the global server decompresses the models and performs aggregation, then compresses them again to send them back. In each communication round, each client $C_i$ uploads its FAM parameters $v^{att}_i$ to the server. Thereafter, the server combines these parameters into a single vector:
\begin{equation}\footnotesize
\label{EQ:agg}
    v^{att}_{\mr{global}} = \frac{1}{N}\sum_{i = 1}^N v^{att}_i.
\end{equation}

\section{Experiments}

\subsection{Datasets}

\mypar{ISIC2019.} ISIC2019 is a skin cancer classification dataset with various characteristics (e.g., age) \cite{tschandl2018ham10000,codella2018skin,combalia2019bcn20000}. To simulate data heterogeneous, the dataset was divided based on the ``anatomy site (AS)'' metadata provided in the original dataset, resulting in seven clients: $C_1$ holds the data whose AS is ``anterior torso'', while $C_2$, $C_3$, $C_4$, $C_5$, $C_6$ and $C_{glo}$ hold the data whose AS is ``head or neck'', ``lower extremity'', ``palms or soles'', ``posterior torso'', ``upper extremity'' and ``Nan'', respectively. 

\mypar{Brain tumor.} The BraTS dataset has two classes, namely high-grade glioma and low-grade glioma \cite{menze2014multimodal,bakas2017advancing,bakas2018identifying,gong2024multi}. It has 273 patients for training, while holds 70 patients for testing. Following \cite{menze2014multimodal}, to simulate heterogeneous data, the training set was divided according to the modalities, resulting in four clients: $C_0$ holds the data whose modality is FLAIR, while $C_1$, $C_2$ and $C_3$ hold the data whose modality is T1 weighted, T1 contrast enhancement (T1-CE), and T2 weighted, respectively. Finally, the original test set (with four modalities) is used for global testing.

\mypar{Prostate cancer.} Similar to BraTS, this dataset has two classes, namely Muscle-Invasive Bladder Cancer (MIBC) and Non-Muscle-IBC (NMIBC) \cite{cao2024multicenter}. It is a multi-center dataset with T2 weighted modality ($C_1$, patients (n) =160; $C_2$, n=48; $C_3$, n=32; $C_4$, n=35), with a total of 279 patients. The acquisition equipment varies from center to center, i.e. $C_1$ has MAGNETOM Skyra, $C_2$ has UMR 780, $C_3$ has Discovery MR750w 3.0T, and $C_4$ has MAGNETOM Verio.

\mypar{ICH.} The RNSA ICH
dataset \cite{flanders2020construction}, which contains five ICH subtypes, is used for experiments. The same pre-processing strategies as in \cite{wu2023fediic} are applied, and images with only a single hemorrhage type are selected. To simulate heterogeneous data, following \cite{wu2023fediic}, Dirichlet distribution is used to divide the training set to \{5,10,15\} clients, while the test set is used for global testing.

For each client, we divide the data into training (60\%), validation (20\%) and test (20\%).

\begin{figure}[!ht]
    \centering
    \includegraphics[width=0.99\linewidth]{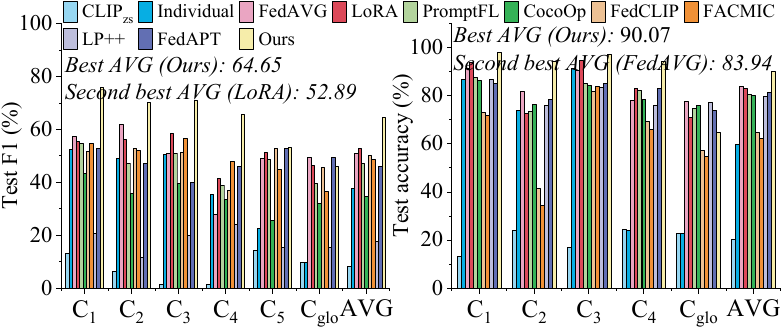}\\
    \caption{Test metrics on ISIC2019 (\textbf{Left}) and BraTS (\textbf{Right}) datasets.}
    \label{fig:ISIC2019_F1_AS}
\end{figure}

\begin{figure}[!ht]
    \centering
    \includegraphics[width=0.9\linewidth]{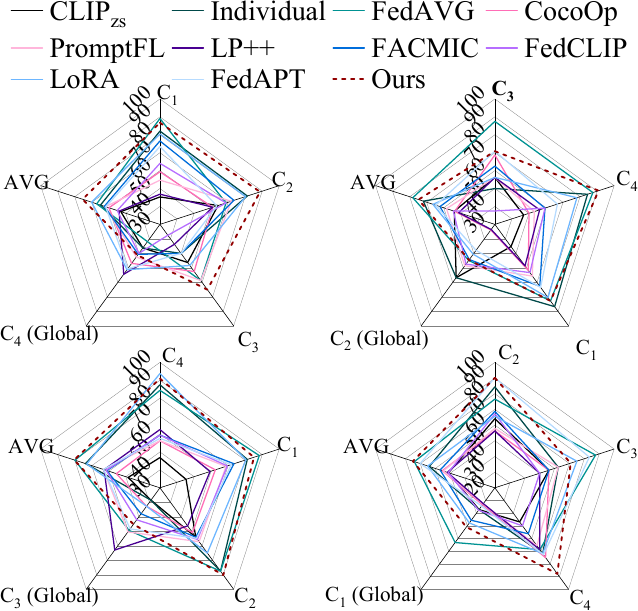}
    \caption{Test accuracy on Prostate dataset.}
    \label{fig:Prostate_ACC}
\end{figure}

\begin{figure}[!ht]
    \centering
    \includegraphics[width=1\linewidth]{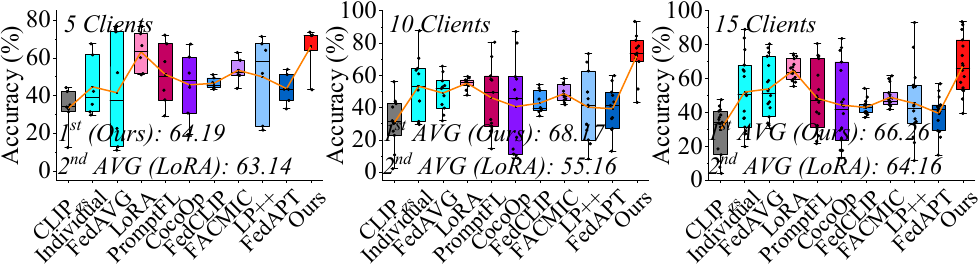}\\ \includegraphics[width=1\linewidth]{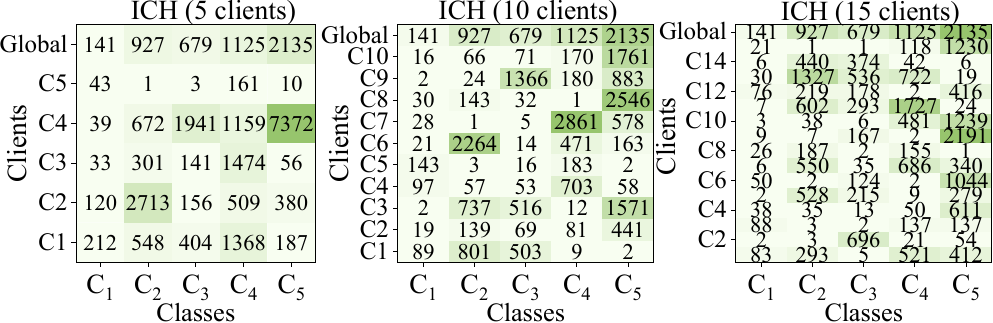}
    \caption{{Test accuracy (\textbf{First row}) and data distributions (i.e., number of samples) (\textbf{Second row}) on ICH dataset. Each box in the first row represents the range of the client accuracies (from min to max), while the square connected by orange line represents the average accuracy.}}
    \label{fig:Boxchart_ICH}
\end{figure}

\begin{table*}[!ht]\footnotesize
    \centering
    \renewcommand{\arraystretch}{0.65}
    \setlength{\tabcolsep}{12.5pt}
    \begin{tabular}{c|cccc|cccc}
    \toprule
         \multirow{2}{*}{Alg.}&\multicolumn{2}{c}{ISIC2019}&\multicolumn{2}{c}{BraTS}&\multicolumn{2}{c}{Prostate}&\multicolumn{2}{c}{ICH} \\
         \cmidrule(l{3pt}r{3pt}){2-6} \cmidrule(l{3pt}r{3pt}){6-9}
         &\multicolumn{4}{c}{Computation (min)}&\multicolumn{4}{|c}{Communication (GB)}  \\

         \midrule
         Individual&35.03 &3.52 &11.72 &75.74 &0 &0 &0 &0 \\
         FedAVG&95.58 &28.19 &98.00 &84.50 &7.569 &3.03 &10.784 &7.07 \\
         LoRA&71.49 &7.201 &17.80 &40.20 &0.259 &0.052 &0.279 &0.174 \\
         PromptFL&50.01 &2.10 &2.05 &61.47 &0.015 &0.0013 &0.0015 &0.018 \\
         CocoOp&86.83 &11.10 &21.80 &90.87 &0.017 &0.0016 &0.0038 &0.018 \\
         FedCLIP&52.75 &3.36 &5.60 &59.75 &0.071 &0.018 &0.054 &0.094 \\
         FACMIC&75.11 &4.64 &15.48 &55.36 &0.094 &0.0181 &0.096 &0.081 \\
         LP++&13.85 &0.47 &4.56 &15.84 &0.0002 &0.000002 &0.000012 &0.000023 \\
         FedAPT&72.495 &4.23 &12.01 &115.30 &0.037 &0.0029 &0.014 &0.038 \\
         Ours&68.56 &2.85 &8.01 &42.15 &0.063 &0.012 &0.054 &0.039 \\ 
         \bottomrule
    \end{tabular}
    \caption{Computation and communication costs for baselines and our approach.}
    \label{tab:computation_comm}
\end{table*}

\begin{table*}[]\footnotesize
    \centering
    \setlength{\tabcolsep}{6pt}
    \renewcommand{\arraystretch}{0.65}
    \begin{tabular}{cccc|cccccc|c|c}
    \toprule
       \multicolumn{4}{c}{Components}&\multicolumn{6}{c}{Clients}&Global&\multirow{2}{*}{AVG}\\
     $\mathcal{L}_{contr}$&$\mathcal{L}_{MLP}$ 
&$\mathcal{L}_{sim}$&Aggregation&$C_1$ & $C_2$& $C_3$& $C_4$&$C_5$&$C_6$& $C_{glo}$& \\
     \midrule
     \checkmark&\ding{56} &\ding{56}&\checkmark&74.28&63.53&78.36&58.82&72.64&74.80&\textbf{81.13}&71.94 \\
     \checkmark&\checkmark &\ding{56}&\checkmark&84.36&69.03&\textbf{84.40}&75.49&77.65&77.60&80.52&78.44\\
     \ding{56}&\checkmark &\ding{56}&\ding{56}&81.23&64.40&80.98&81.37&76.69&72.60&17.17&67.77\\
     \checkmark&\checkmark &\checkmark&\ding{56}&81.67&69.75&82.80&\textbf{84.31}&\textbf{80.35}&76.80&17.17&70.41\\
    \checkmark&\checkmark &\checkmark&\checkmark&\textbf{84.45}&\textbf{71.92}&82.69&\textbf{84.31}&79.00&\textbf{79.40}&81.01&\textbf{80.4}\\
     \bottomrule
    \end{tabular}
    \caption{{Ablations on each component in our approach. \textbf{Bold} represents the best.}}
    \label{tab:ablations}
\end{table*}

\mypar{Implementation details.} Pretrained ViT-B/32 provided by OpenAI \cite{radford2021learning} is used as the CLIP backbone with a batch sise of 32. We use the AdamW optimizer to fine-tune the parameters of both the FAM and the MLP. To provide a stable learning procedure as suggested in \cite{wu2024facmic}, the initial learning rate of the FAM is set to $5 \times 10^{-5}$ for Prostate and BraTS, while $5 \times 10^{-4}$ for ISIC2019. For MLP, it is set to $1 \times 10^{-3}$ (ISIC2019) and $1 \times 10^{-4}$ (Prostate and BraTS). The $\lambda$ and $T$ are set to 0.04 and 2 for all datasets, respectively. The optimization process uses a weight decay of $0.02$ and beta parameters of $(0.99, 0.98)$. The exponential learning rate scheduler is used with a gamma of 0.97 for local training. We set the communication round to 100 (ISIC2019) and 50 (BraTS, Prostate and ICH), while the local training epoch is set to one. We select one vanilla FL technique (FedAVG), and seven PEFT based federated approaches, namely FedCLIP, PromptFL, CocoOp with FedAVG \cite{zhou2022conditional}, LP++ with FedAVG \cite{huang2024lp++}, LoRA with FedAVG (rank is set to three) \cite{zanella2024low}, FedAPT and FACMIC. For CocoOp, only the parameters of the prompt learner are aggregated, whereas for LP++, only the linear probes parameters are aggregated. For non-federated methods, the Individual without aggregation and CLIP$_{zs}$ are selected. The random seed was set to 0 to eliminate the impact of different seeds. All experiments are based on the Windows 11 operating system, and feature an Intel 13900KF CPU with 128 GB of RAM and an RTX 4090 GPU. The FAM has $\sim 5\times10^5$ parameters.

\mypar{Results.} Table \ref{tab:ISIC2019}, Figure \ref{fig:ISIC2019_F1_AS}, Figure \ref{fig:Prostate_ACC}, and Figure \ref{fig:Boxchart_ICH} show the test accuracy (ACC) and F1 score for the ISIC2019, BraTS, Prostate and ICH datasets. We highlight the following points: 1) For all medical datasets, the original pre-trained CLIP shows poor generalization ability (e.g., $24.1\%$ AVG test ACC on ISIC2019), consistent with the results described in \cite{huix2024natural}. This suggests that the original CLIP has limited domain knowledge. 2) The use of FL techniques can improve the performance of CLIP on medical datasets (e.g., using FedAVG provides 72.12\% AVG ACC on ISIC2019). However, they still face challenges such as class imbalance, as it shows a lower F1 score (e.g. 51. 03\% AVG on ISIC2019), indicating overfitting in unbalanced data. Furthermore, for the clients with sufficient data (e.g., $C_1$ in Prostate), the learned knowledge can improve the performance of FAM and MLP in other clients (e.g., $86.72\%$ and $88.73\%$ ACC in $C_2$ and $C_3$, client:\{$C_1$,$C_2$,$C_3$\}, global: $C_4$). In addition, for the clients with limited samples (e.g., $C_3$ in Prostate), the usefulness of FAM is weak and it will negatively degrade the feature representations, thereby reducing the performance of local clients (e.g., $37.5\%$ ACC on $C_3$ with FedCLIP). However, the proposed approach improves performance on $C_3$ to 75\% by training a FAM with a local MLP mutually. A similar situation can be found for $C_4$ in the following setting: client:\{$C_2$,$C_3$,$C_4$\}, global: $C_1$. 3) Considering the average performance, the proposed approach yields the highest AVG metrics (e.g., 80.4\% ACC and 64.65\% F1 score on ISIC2019, 90.07\% ACC on BraTS) compared to other baselines. Futhermore, FedAVG shows large differences on ICH dataset (e.g., larger box size in Figure \ref{fig:Boxchart_ICH}), while PEFT methods such as FedCLIP provide smaller discrepancies. However, these approaches show considerable performance degradation as the number of clients increases (e.g., $\sim5\%$ ACC drop with FACMIC from 5 to 10 clients). Unlike others, our method provides stable test metrics despite client settings (e.g., consistently higher than 66\% AVG), suggesting that the introduction of masked MLP and KL based regularizations is a robust solution for a large-scale client setting. 4) Regarding the global generalization ability, the proposed FAM yields feasible ACC on ISIC2019 (81.01\% ACC), while providing lower ACC on BraTS (64.66\%) compared to other PEFT methods such as FedAPT (74.04\%). This suggests that the potential of FAM is limited where the feature shifts are considerablely large (e.g., different modalities). We argue that the pre-trained CLIP encoders has limited prior knowledge, thus simply using a FAM on the global site leads to $\sim$4\% lower ACC on ISIC2019 dataset compared to prompt-based methods such as FedAPT, while for local clients, benefited by the MLP, it learns robust feature patterns, thus providing the best test metric.

\begin{figure}[!ht]
    \centering
    \includegraphics[width=0.9\linewidth]{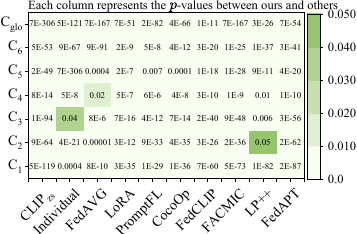}
    \caption{{The $p$-value between our model and the other baselines on the ISIC 2019 dataset using Wilcoxon signed-rank test. These values are based on the test set.}}
    \label{fig:P_Values}
\end{figure}

\begin{figure*}[!ht]
    \centering
    \includegraphics[width=0.19\linewidth]{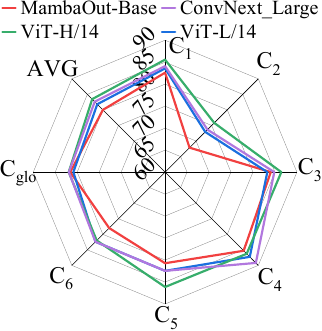}
    \includegraphics[width=0.225\linewidth]{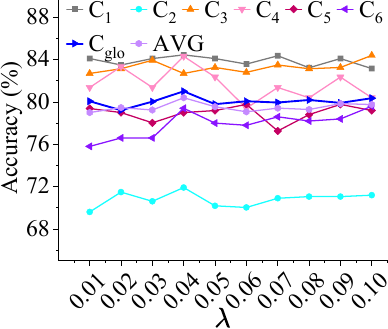} 
    \includegraphics[width=0.225\linewidth]{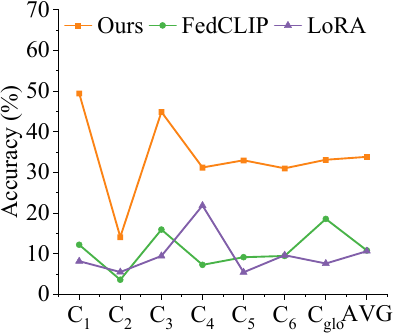}
    \includegraphics[width=0.225\linewidth]{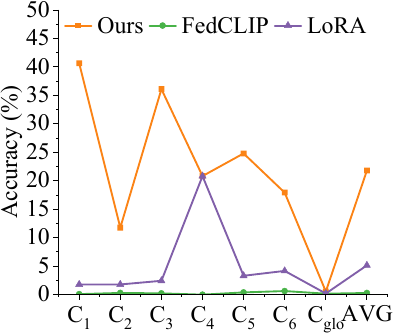}
    \caption{{Test accuracy on ISIC2019 dataset with varying backbones (\textbf{Left}), $\lambda$ values (\textbf{2-column}), FGSM (\textbf{3-coumn}) and PGD (\textbf{Right}).}}
    \label{fig:ISIC2019_ablation_attacks}
\end{figure*}

\mypar{Computation and communication overhead.} Table \ref{tab:computation_comm} reports the total computation time and communication cost for the baselines and the proposed approach to achieve the best validation metrics. For Prostate, only one client setting is considered (client:\{C$_1$,C$_2$,C$_3$\}, global: \{C$_4$\}), while for ICH, the results are based on 10 clients. As illustrated, FedAVG has a large computational and communication overhead (e.g., 95.58 min, 7.569 GB on ISIC2019), while the proposed method provides a feasible resource cost (e.g., 120$\times$ faster than FedAVG). Furthermore, our method yields comparable computational and communication load compared to baselines such as FedAPT on ISIC2019 (68.56 min, 0.063 GB vs. 72.495 min, 0.037 GB), and FedCLIP on BraTS (2.85 min, 0.012 GB vs. 3.36 min, 0.018 GB). These results highlight the potential of the proposed designs for resource-efficient FL frameworks.

\mypar{Ablations on components.} We validate the usefulness of each component in the proposed approach in the ISIC2019 data set. As illustrated in Table \ref{tab:ablations}, introducing $\mathcal{L}_{MLP}$ improves the overall performance in local clients (e.g., $\sim 7\%$ ACC improvement), while adding $\mathcal{L}_{sim}$ provides 1.32\% average ACC improvement compared to without $\mathcal{L}_{sim}$. We note that for certain clients, such as $C_1$, the use of $\mathcal{L}_{sim}$ slightly reduces the performance (e.g., 1.21\%). This suggests that FAM can reduce the learned knowledge of the local MLP in class imbalance situation. Furthermore, the usefulness of $L_{sim}$ on the global site performance is limited since it lacks a private local MLP (i.e., it can not provide ensemble prediction). {Overall, the use of all these components leads to the best AVG (80.4\%).}

\noindent\textbf{Statistical Significance Analysis.} Figure \ref{fig:P_Values} shows the $p$-value (measured with test set, significance level equals 0.05) between ours and other baselines on the ISIC2019 dataset using Wilcoxon signed-rank test \cite{rey2011wilcoxon}. As illustrated, the proposed method demonstrates significant performance improvements over other approaches (e.g., $p<10^{-11}$ on $C_2$ compared to LoRA).

\begin{figure}[!ht]
    \centering
    \includegraphics[width=0.37\linewidth]{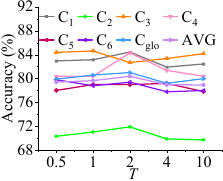} \includegraphics[width=0.38\linewidth]{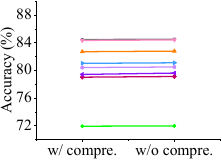}
    \caption{Test accuracy of the proposed method on ISIC2019 dataset with different $T$ values (\textbf{Left}) and model compression (\textbf{Right}).}
    \label{fig:T_Values}
\end{figure}

\mypar{Influence of $T$.} We explored the performance on the ISIC2019 dataset with different $T$ values used in Eq. \ref{EQ:T}. As illustrated in Figure \ref{fig:T_Values}, smaller $T$ leads to a decrease in performance (e.g., an ACC of 78.03\% on $C_5$), while larger $T$ such as 10 results in a lower AVG of 78.94\%.

\begin{figure*}[ht]
    \centering
    \includegraphics[width=0.138\linewidth]{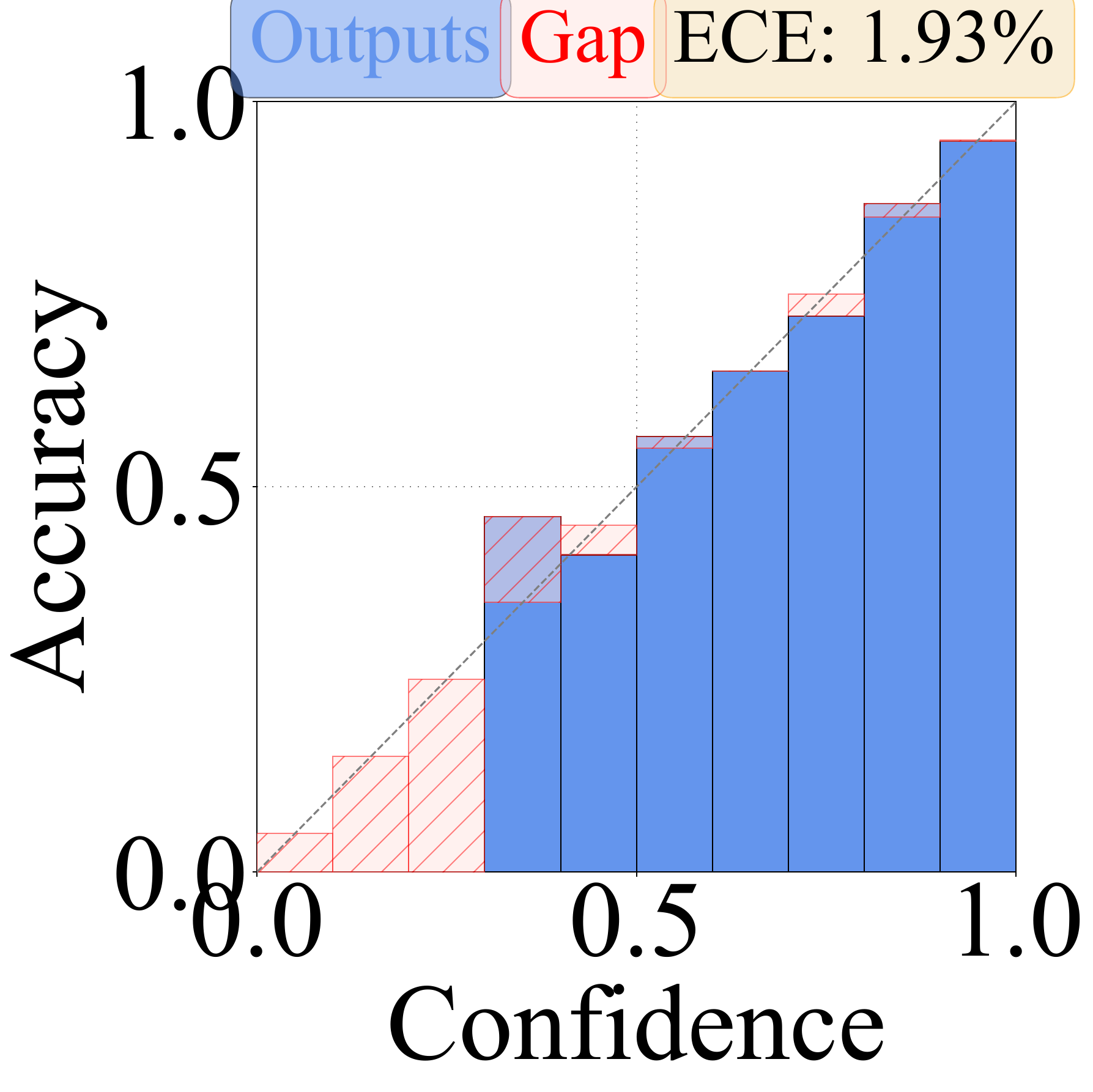} \includegraphics[width=0.138\linewidth]{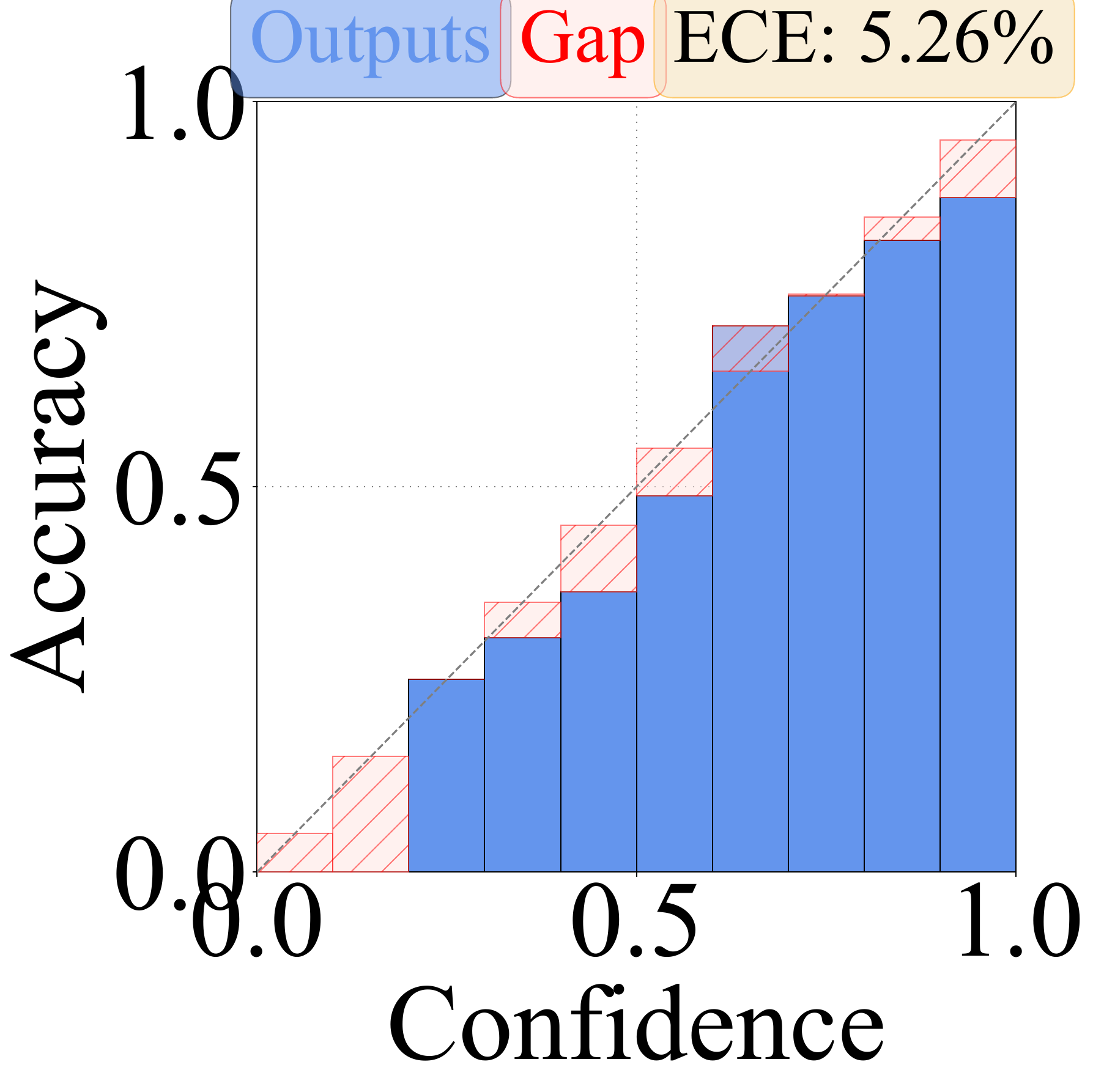} \includegraphics[width=0.138\linewidth]{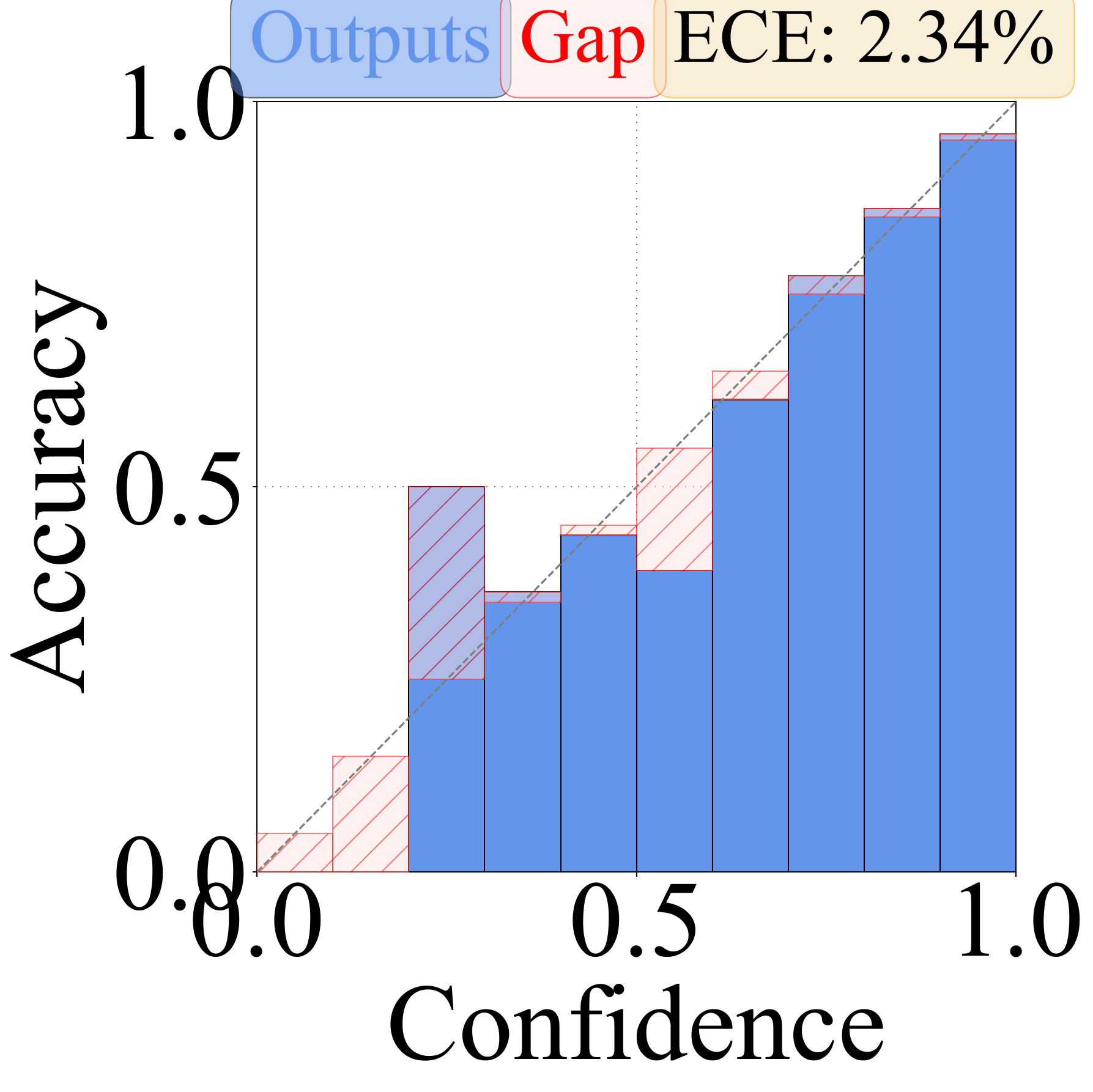} \includegraphics[width=0.138\linewidth]{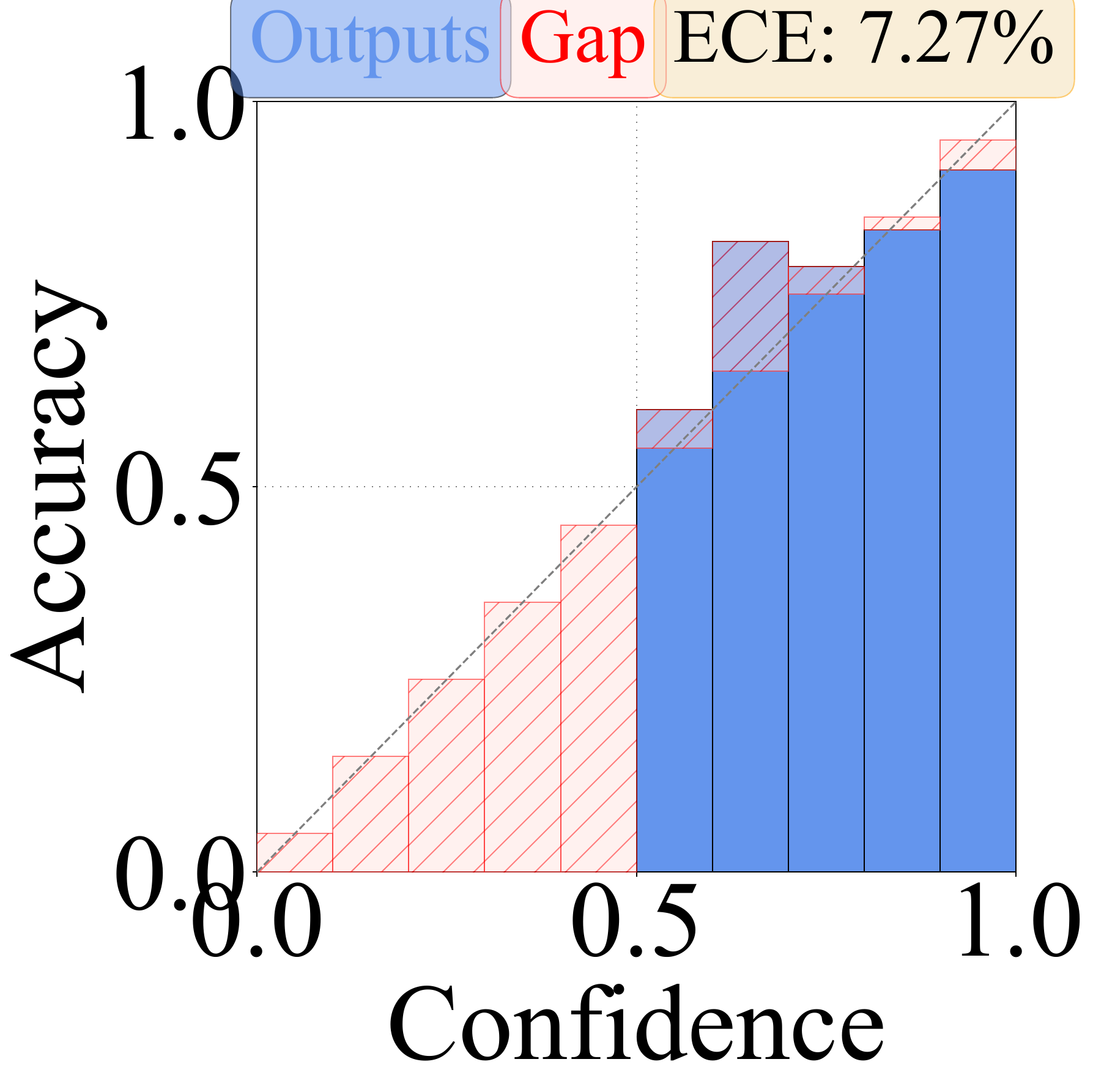} \includegraphics[width=0.138\linewidth]{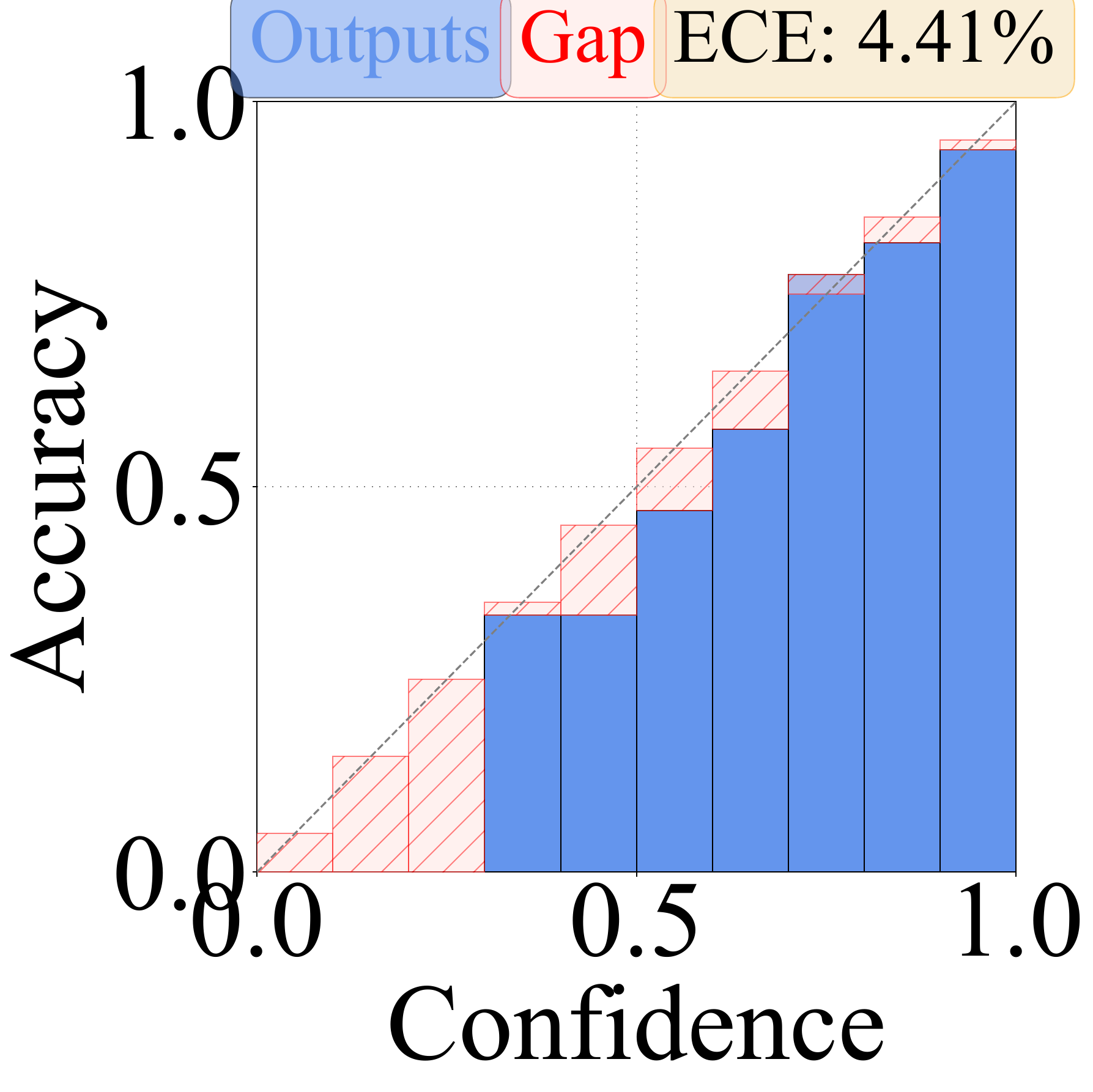} \includegraphics[width=0.138\linewidth]{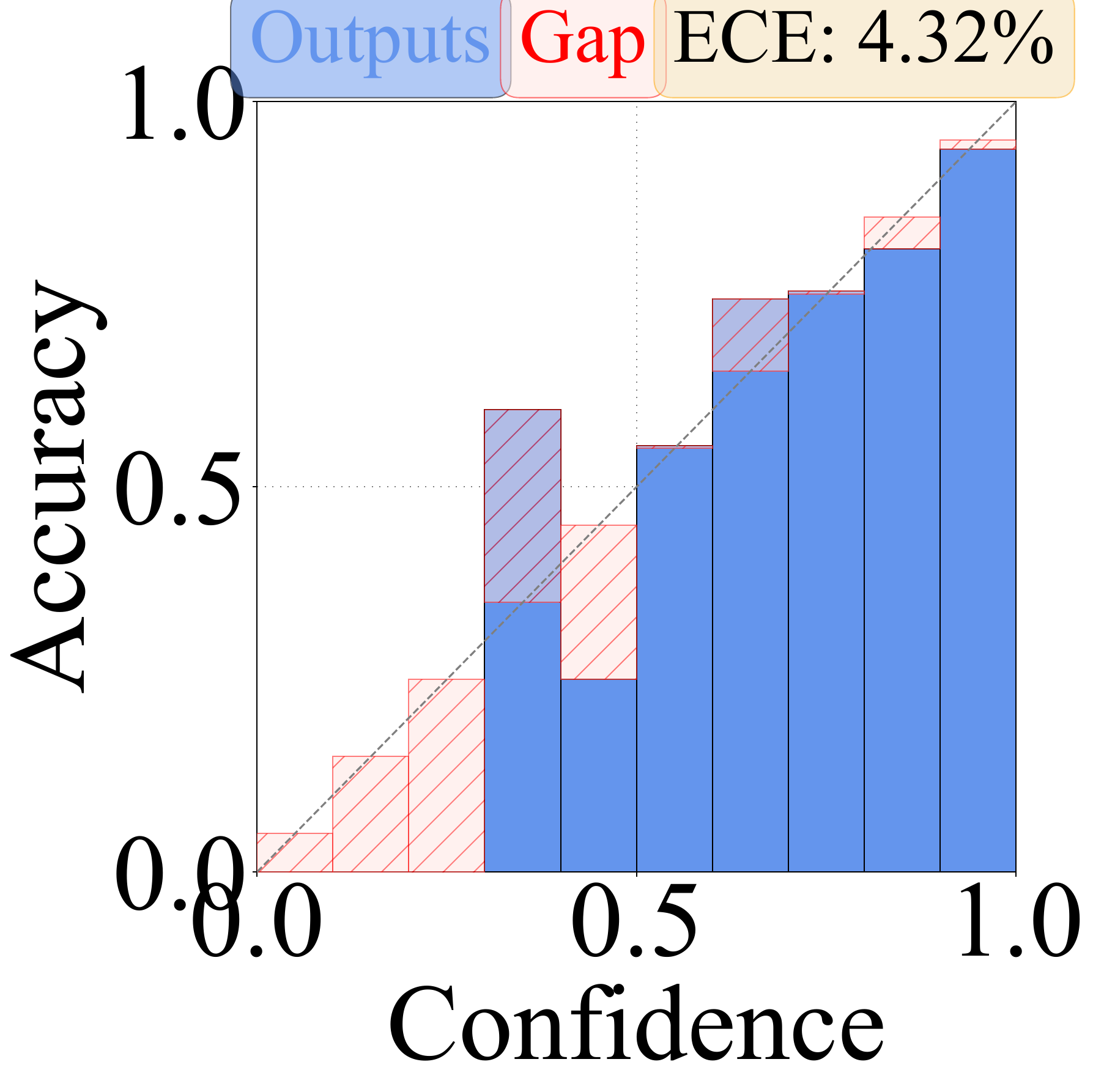} \includegraphics[width=0.138\linewidth]{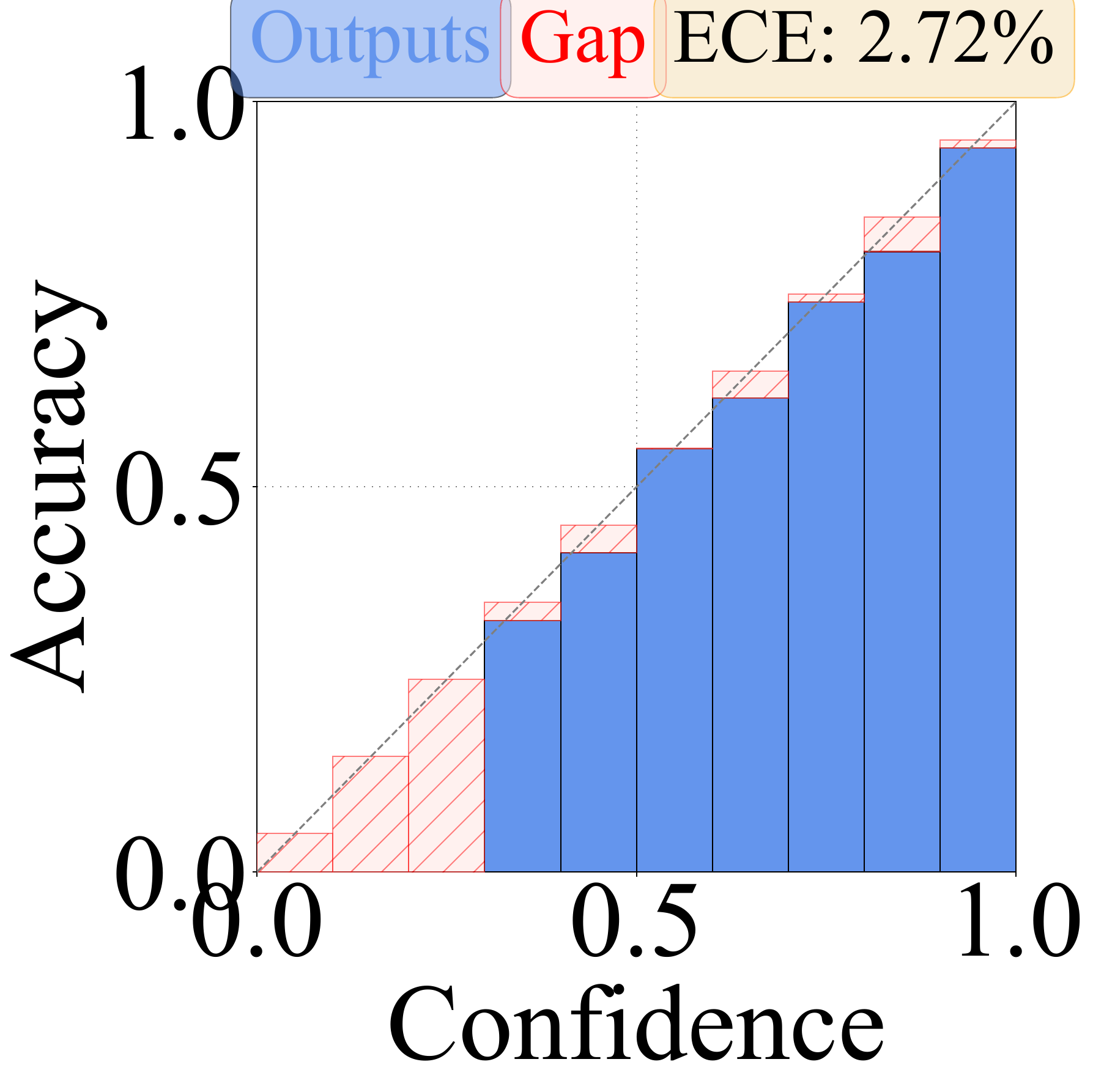}\\
    \includegraphics[width=0.138\linewidth]{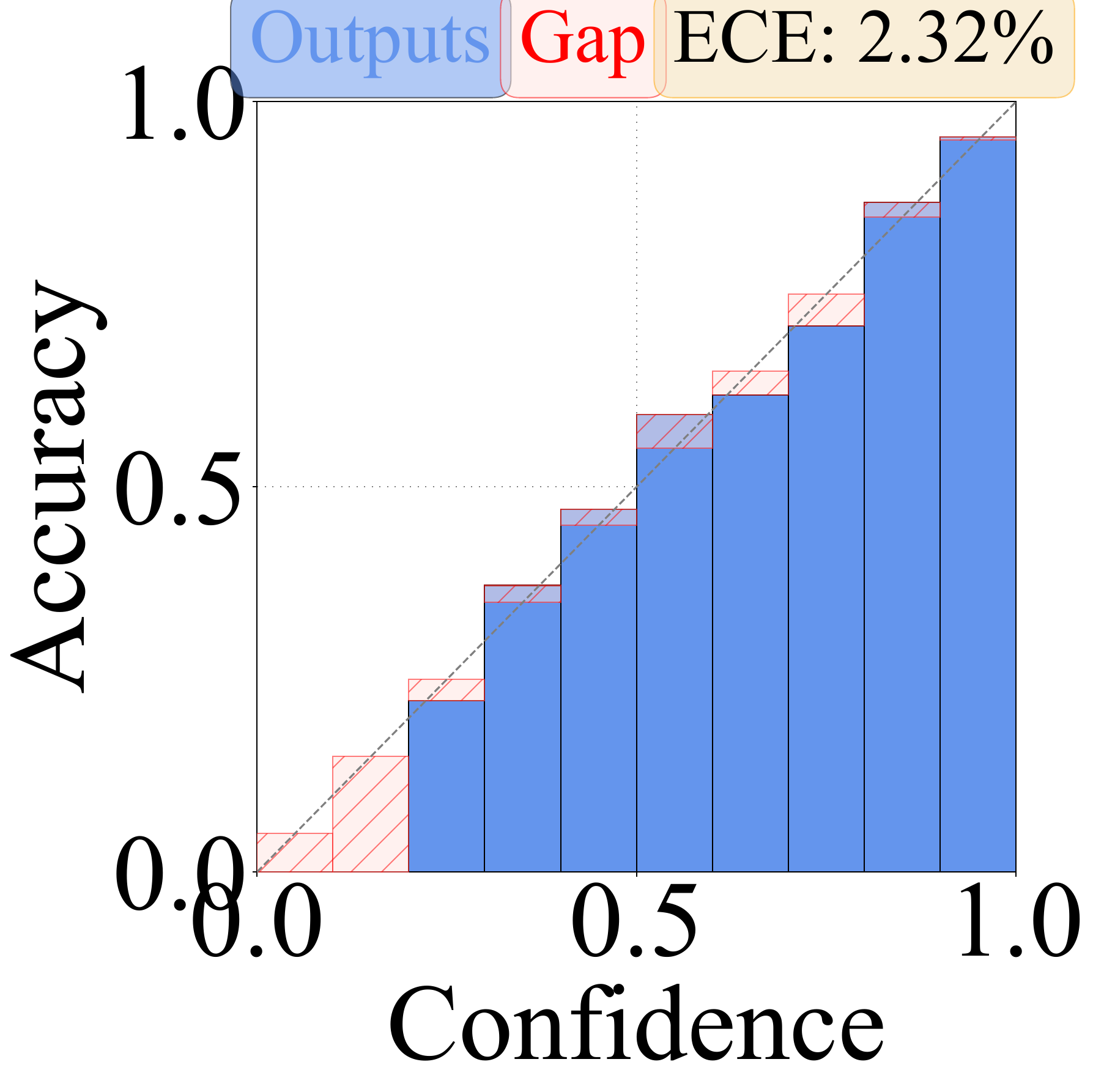}
    \includegraphics[width=0.138\linewidth]{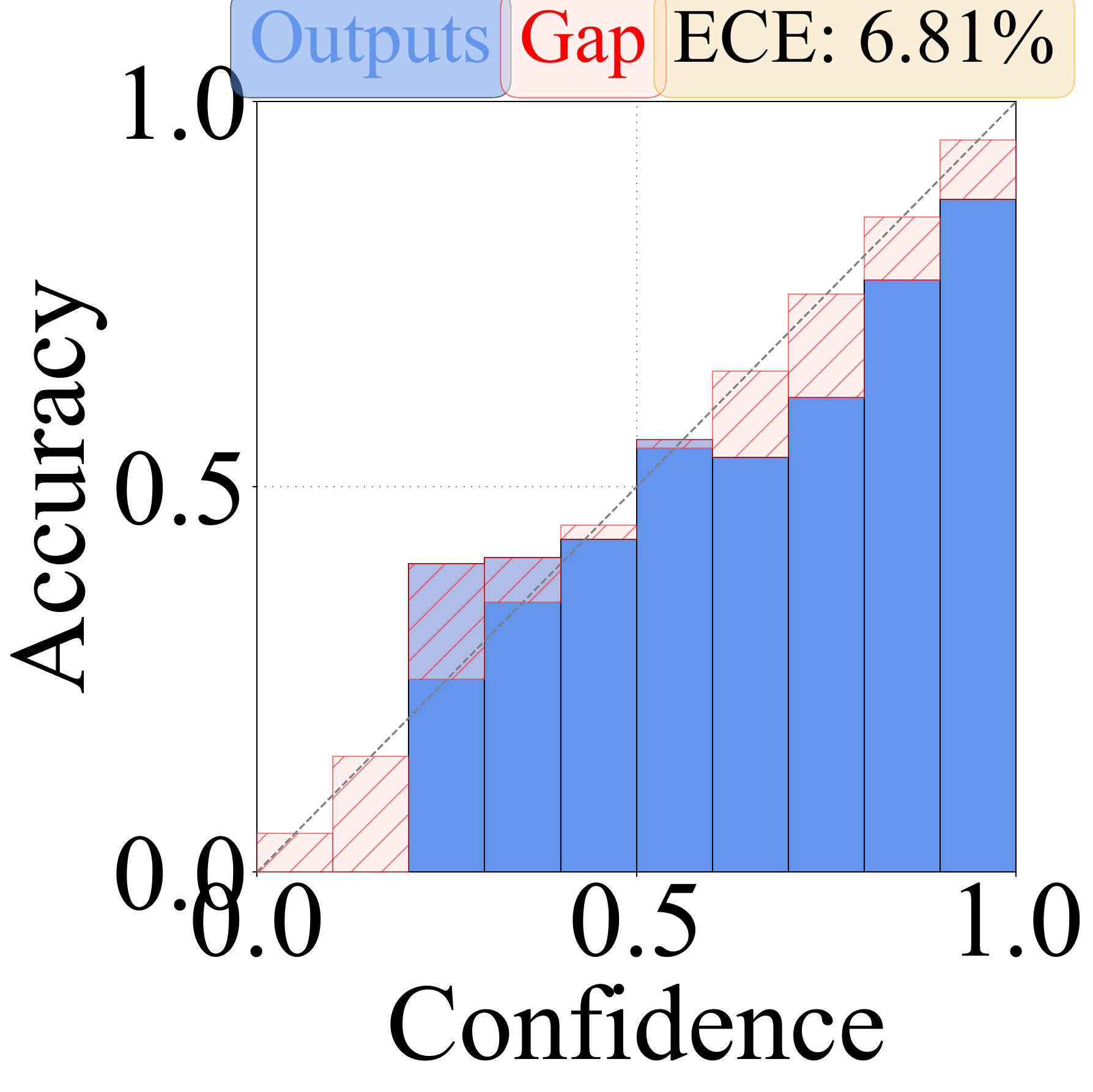}
    \includegraphics[width=0.138\linewidth]{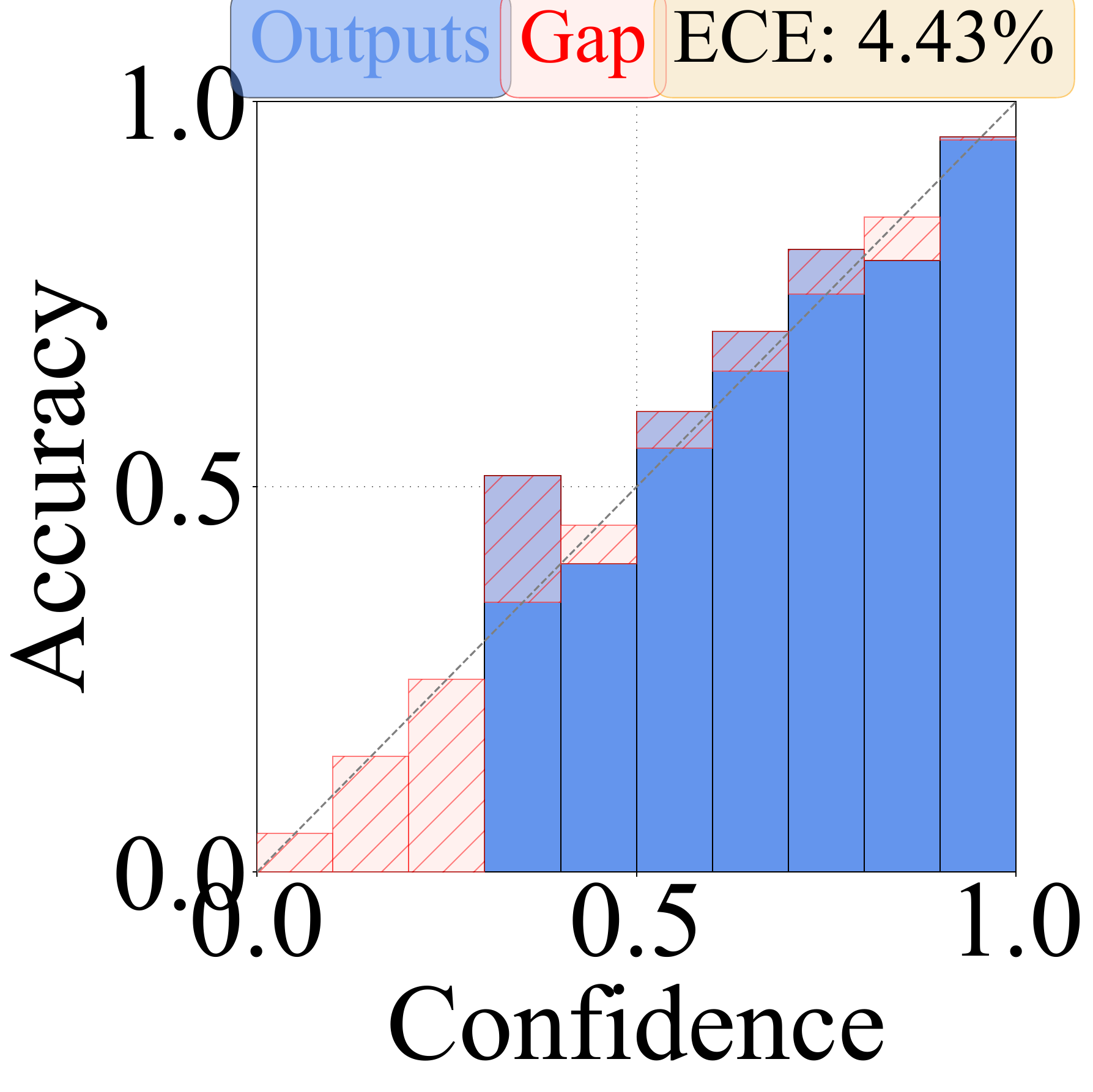}
    \includegraphics[width=0.138\linewidth]{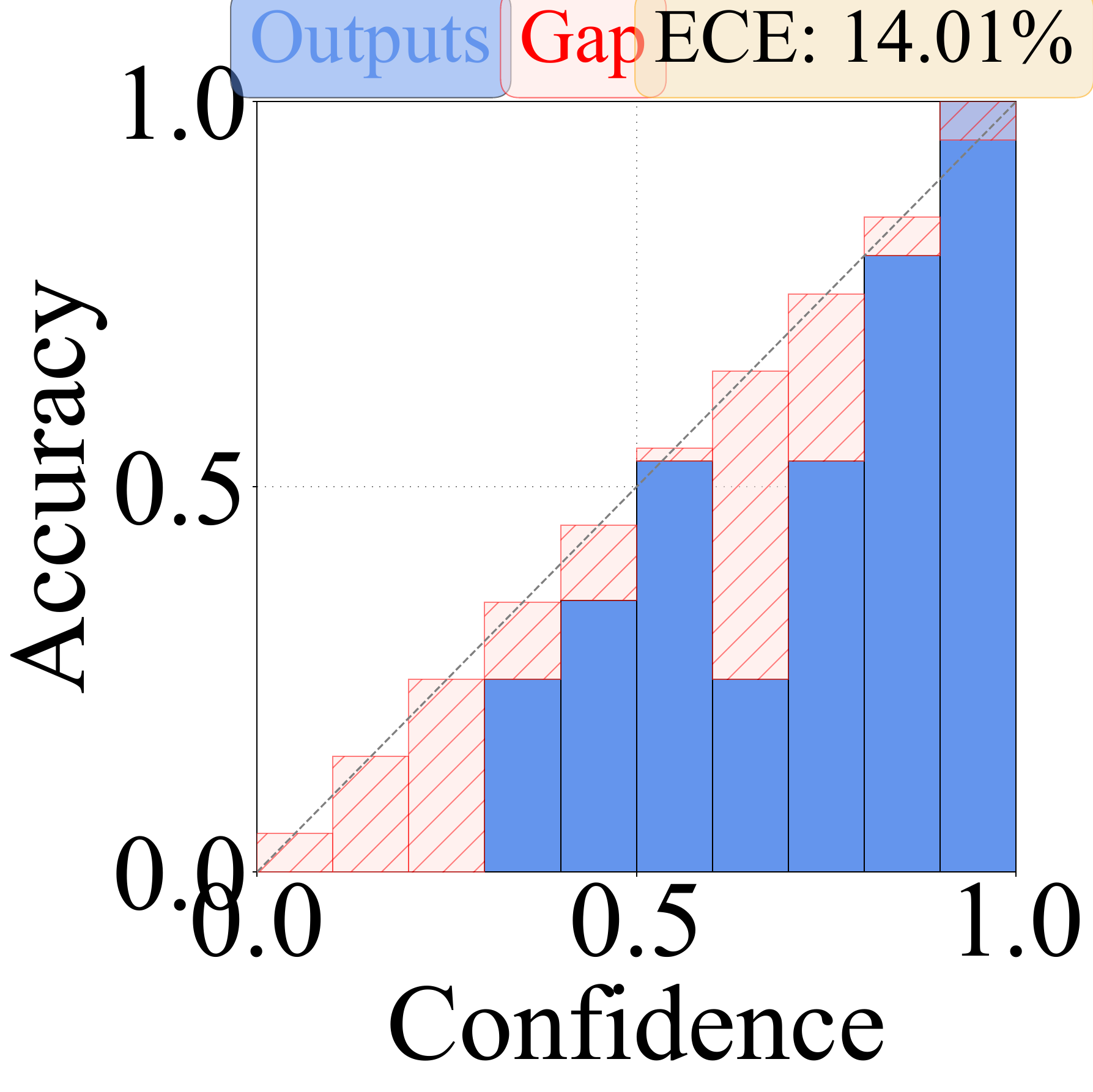}
    \includegraphics[width=0.138\linewidth]{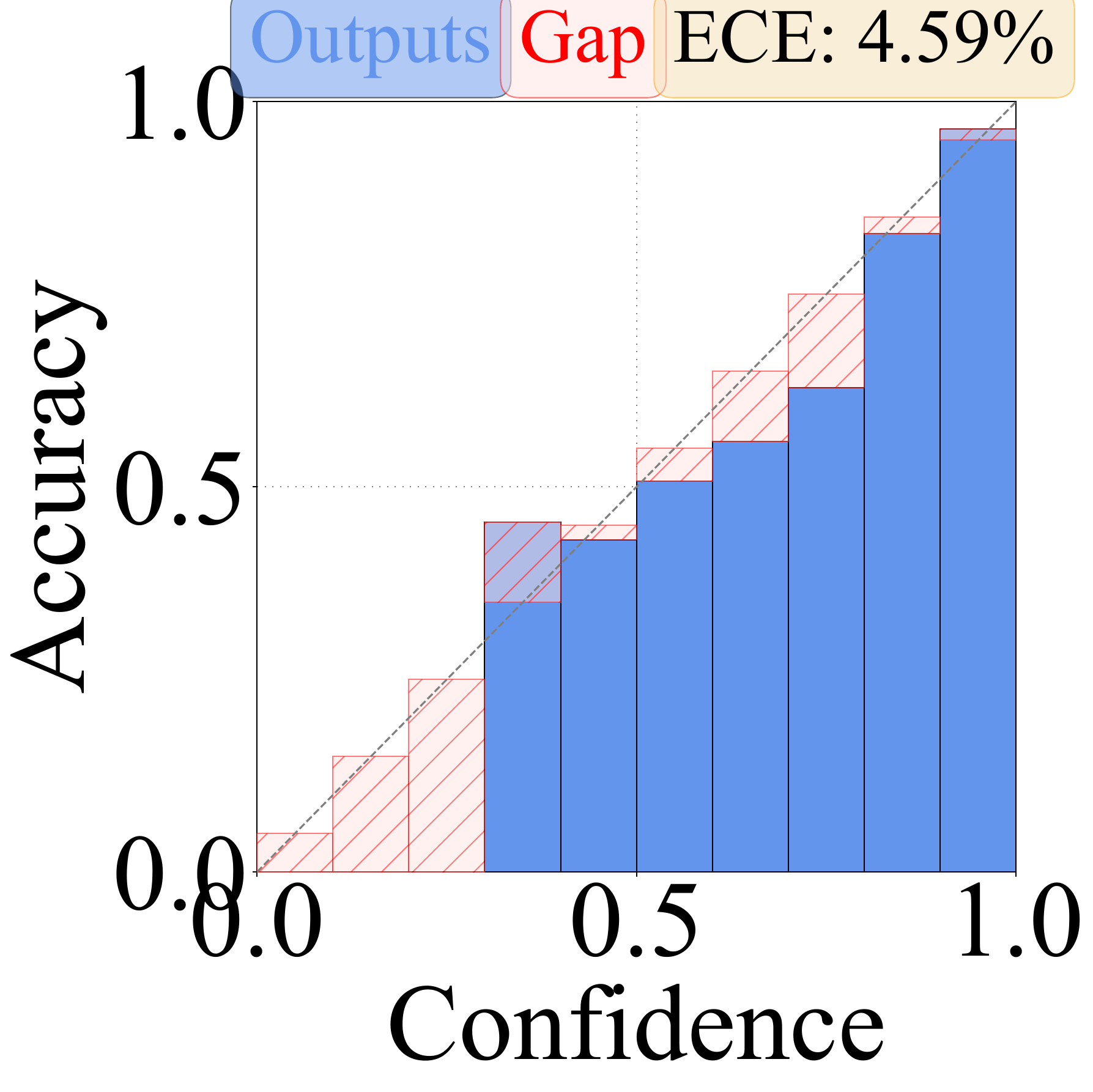}
    \includegraphics[width=0.138\linewidth]{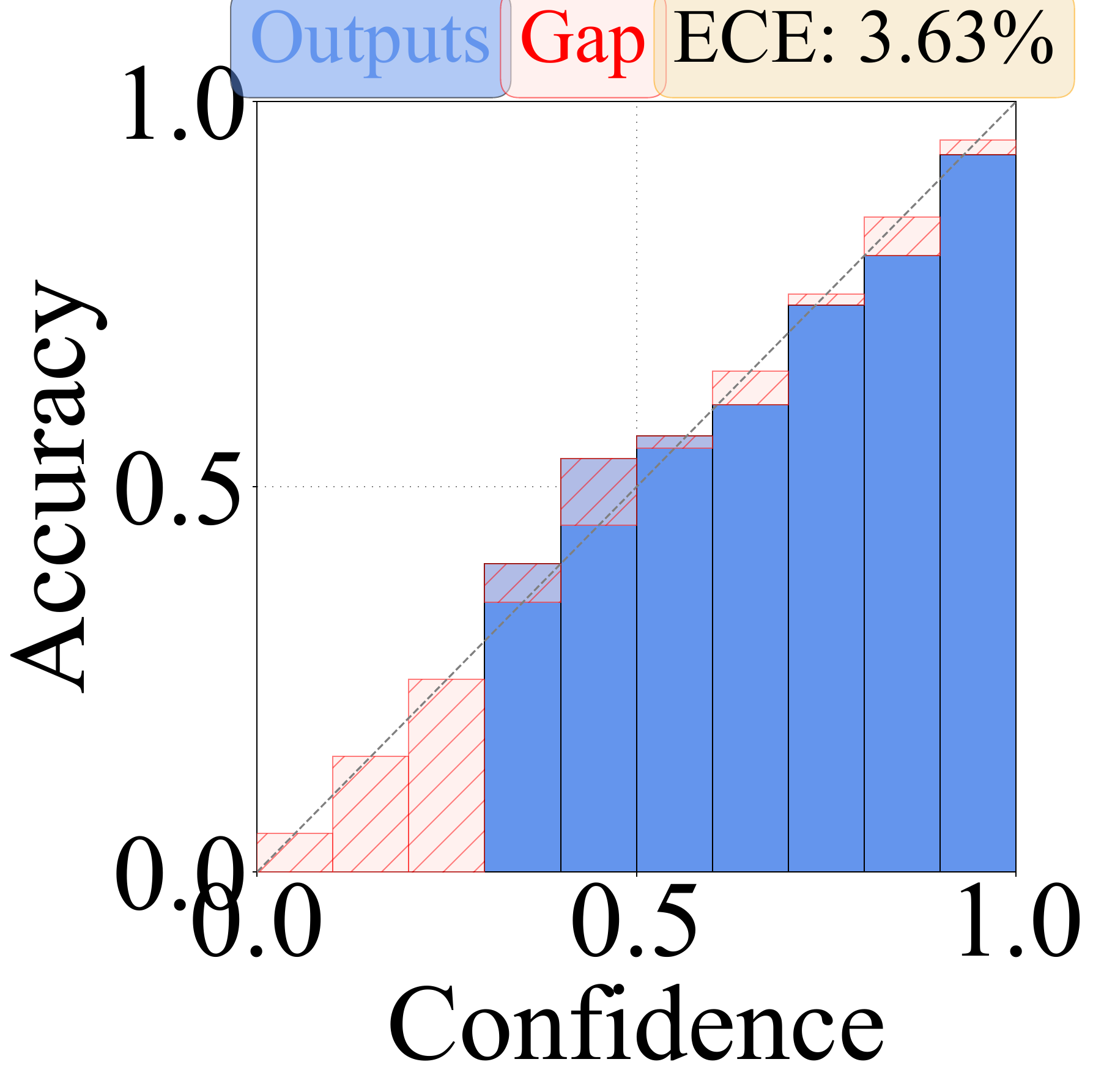} \includegraphics[width=0.138\linewidth]{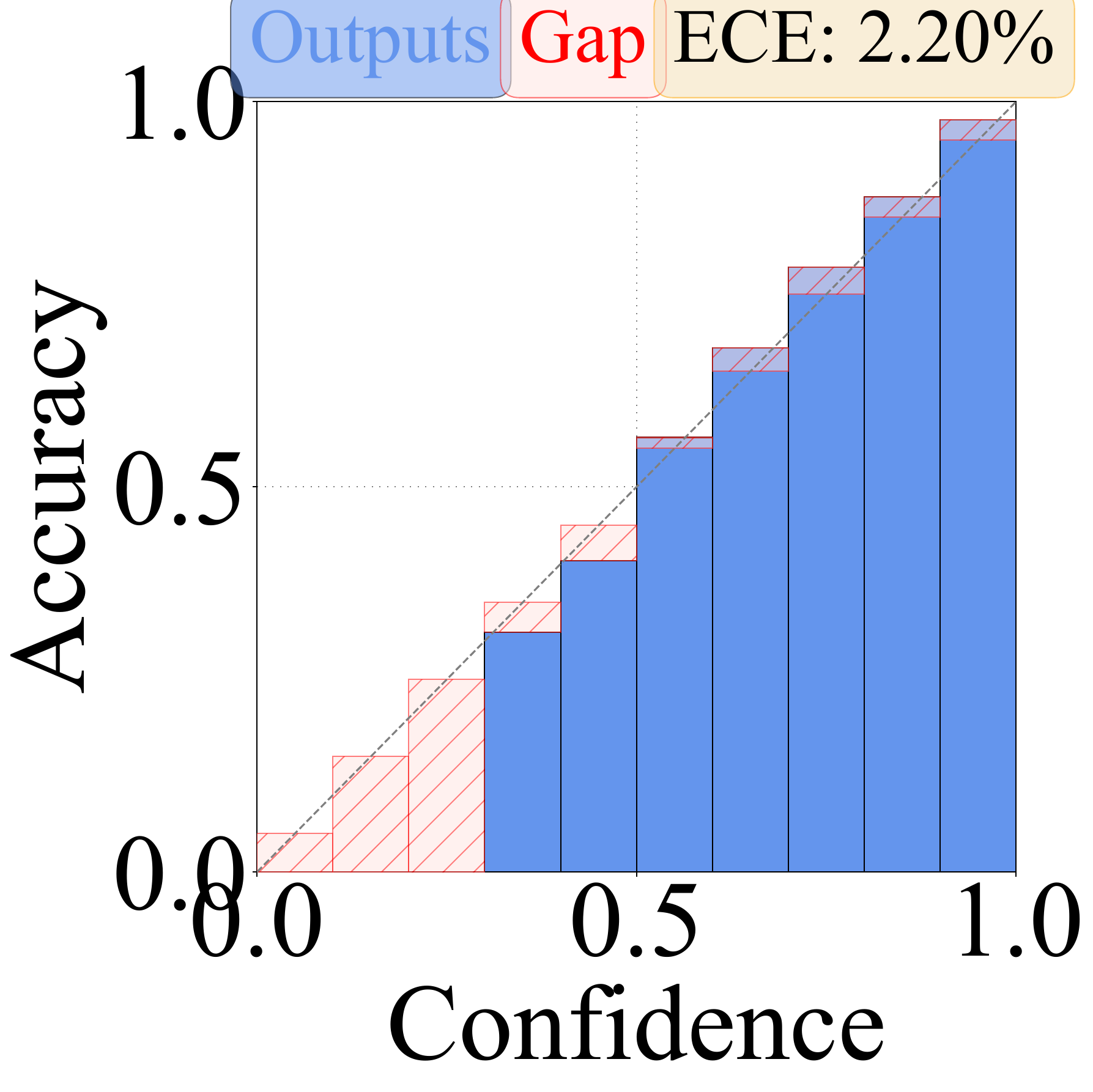} \\
    \includegraphics[width=0.138\linewidth]{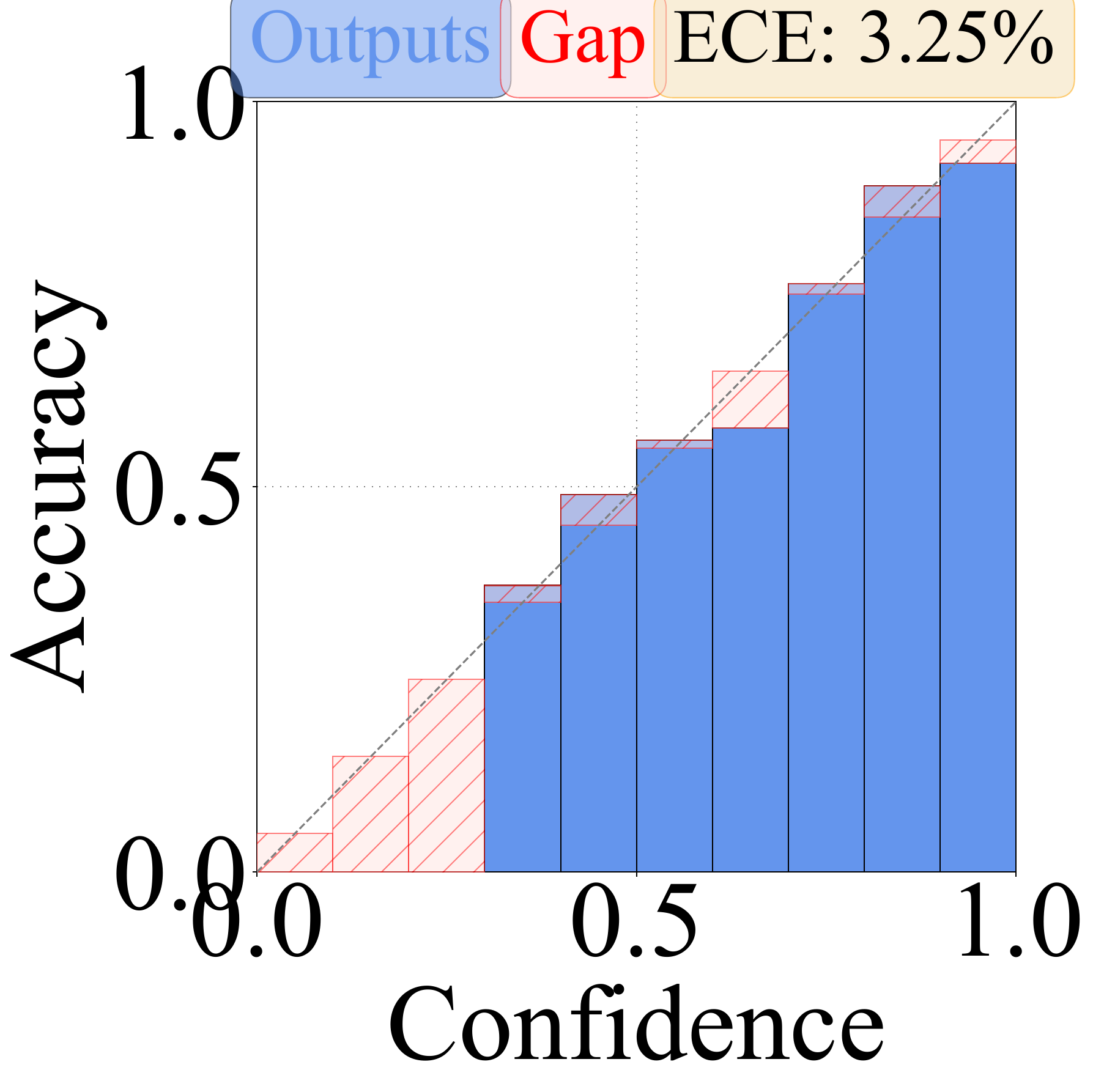}
    \includegraphics[width=0.138\linewidth]{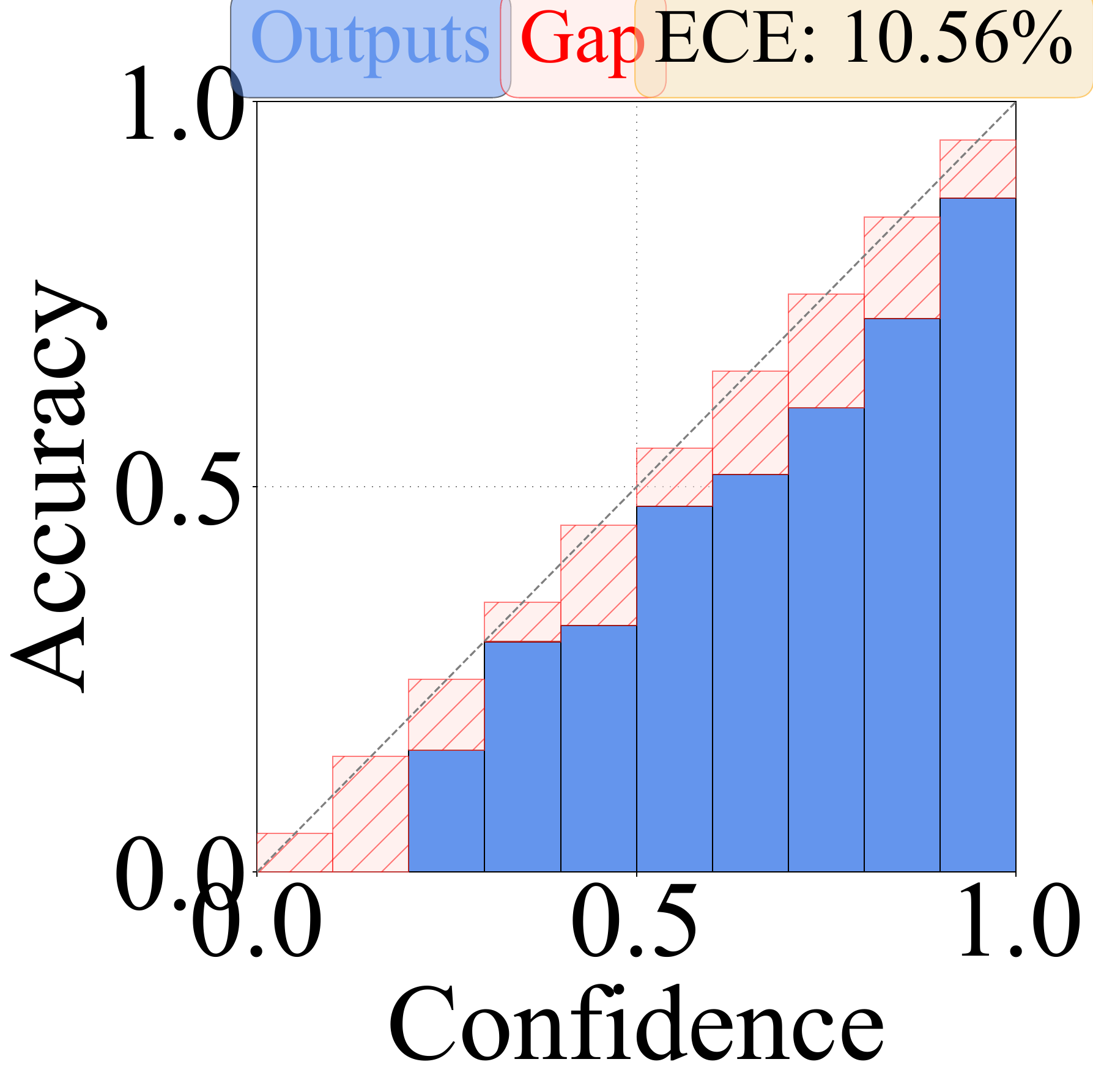}
    \includegraphics[width=0.138\linewidth]{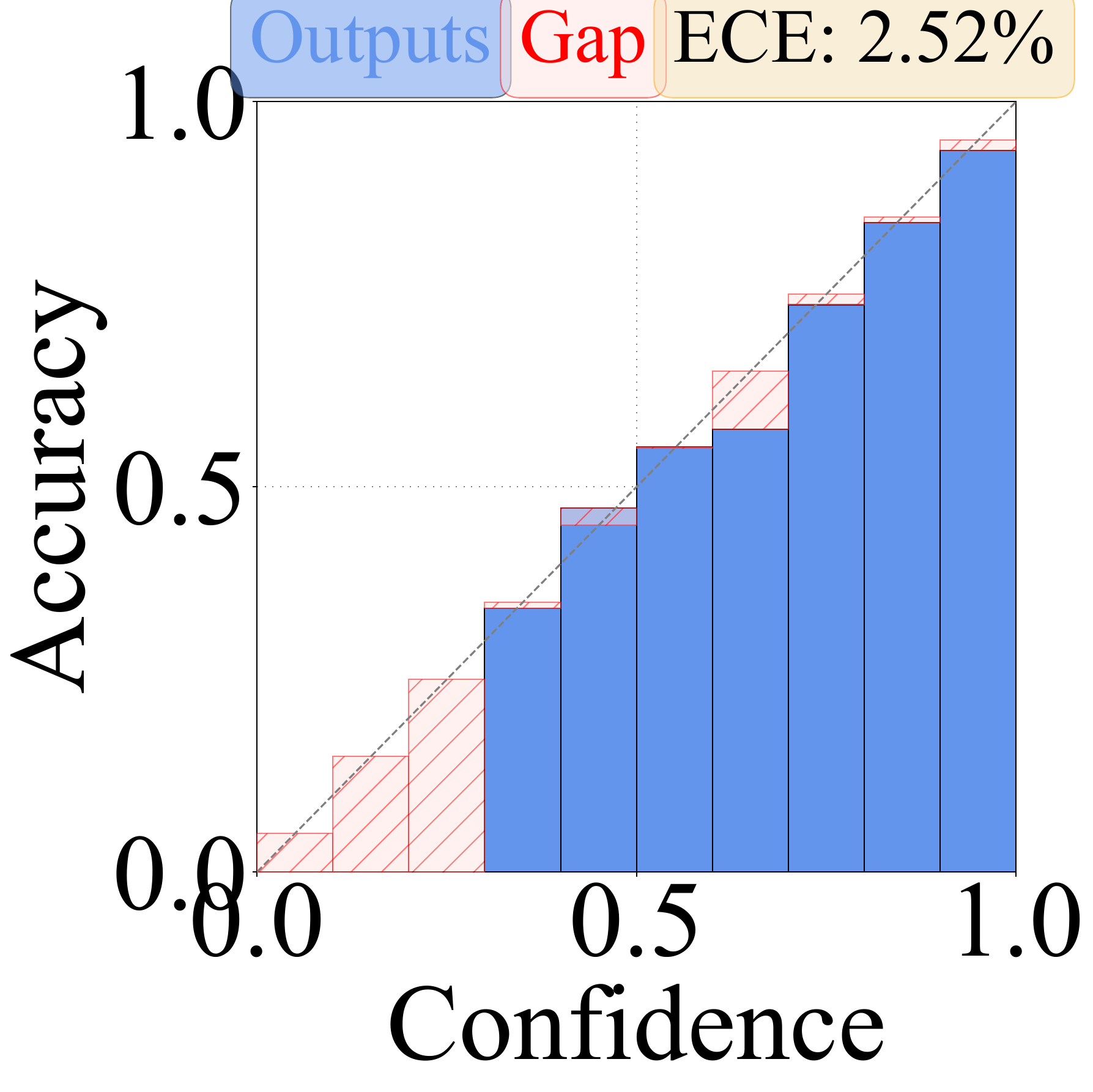}
    \includegraphics[width=0.138\linewidth]{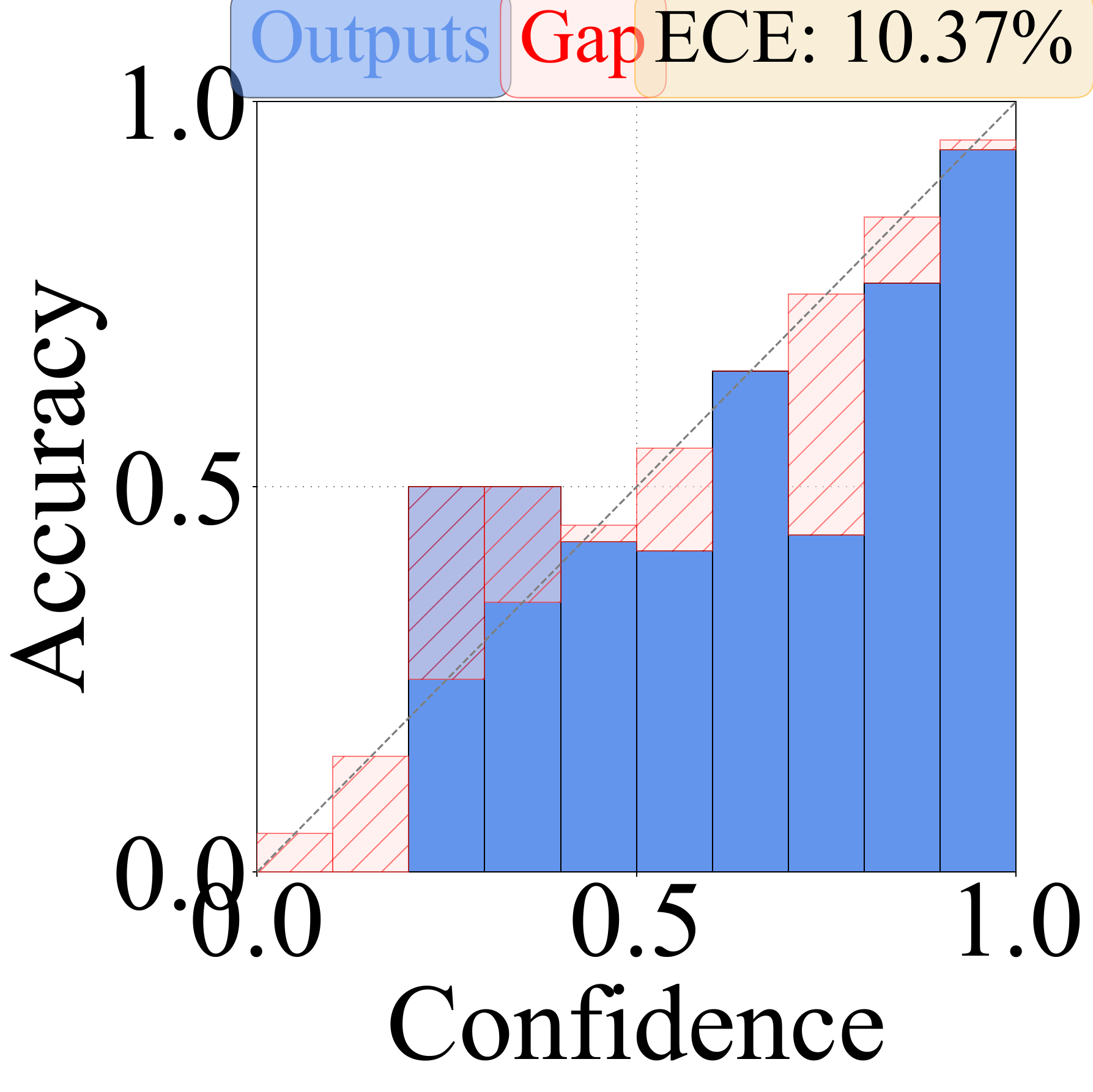}
    \includegraphics[width=0.138\linewidth]{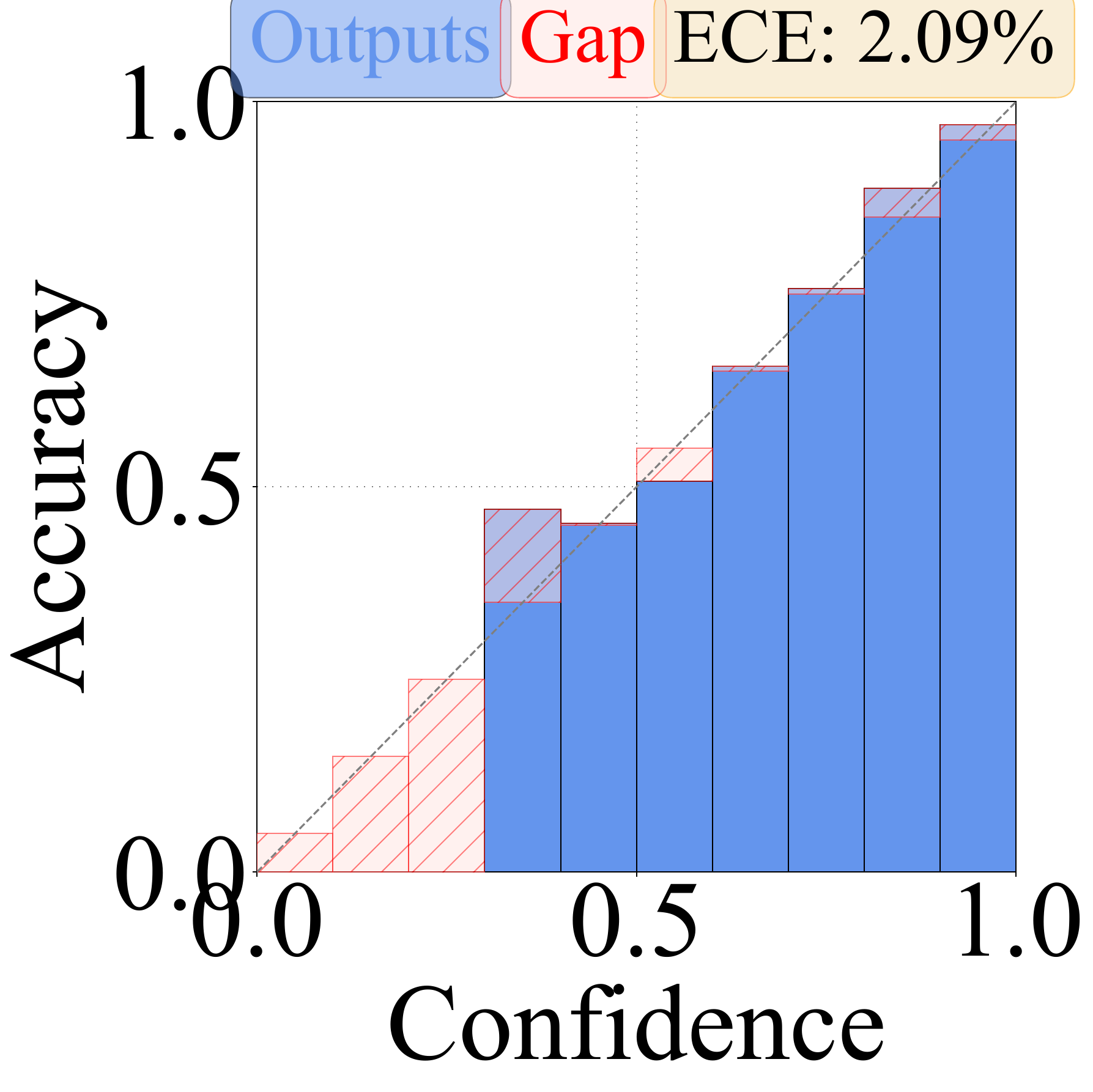}
    \includegraphics[width=0.138\linewidth]{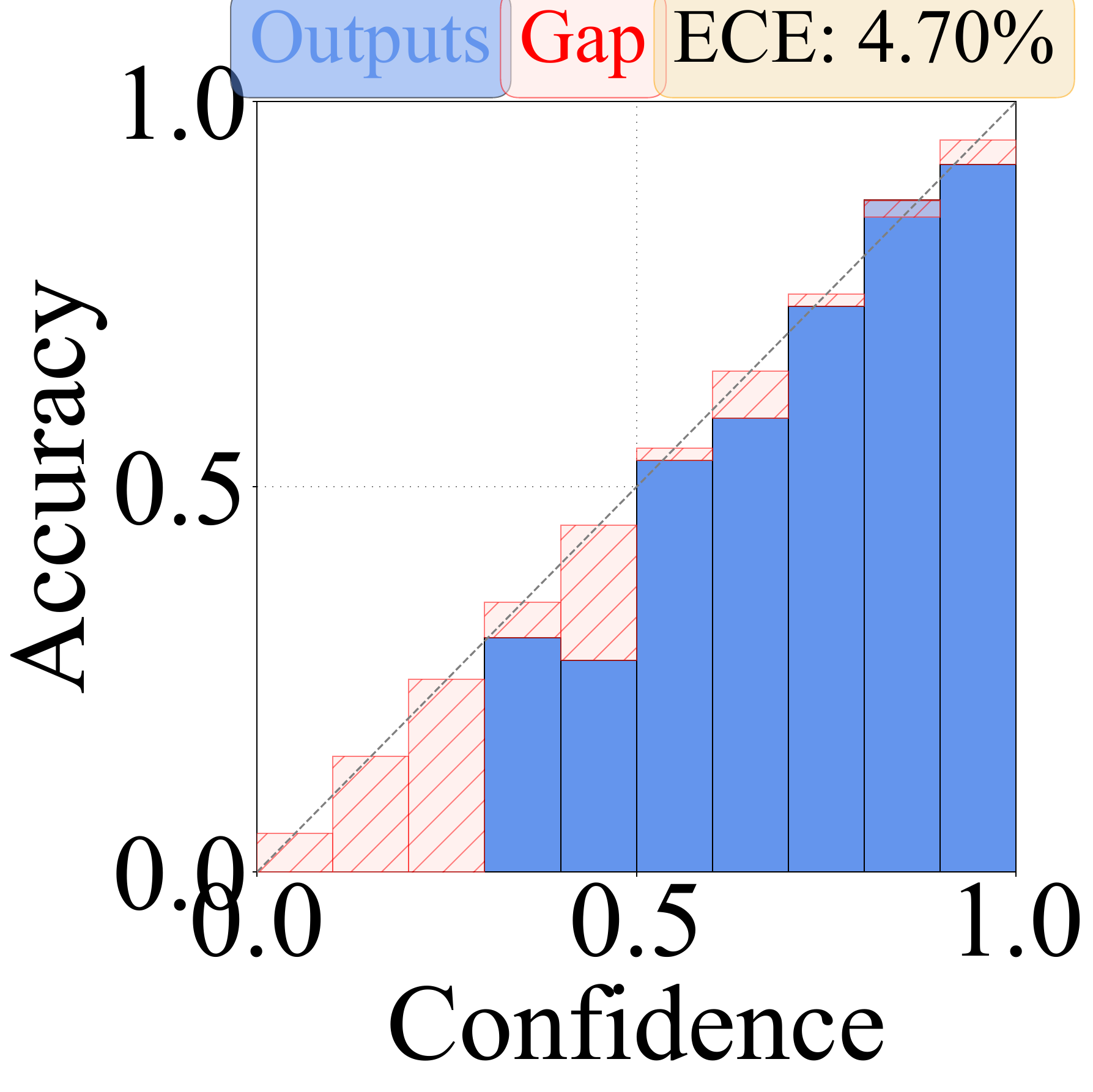} \includegraphics[width=0.138\linewidth]{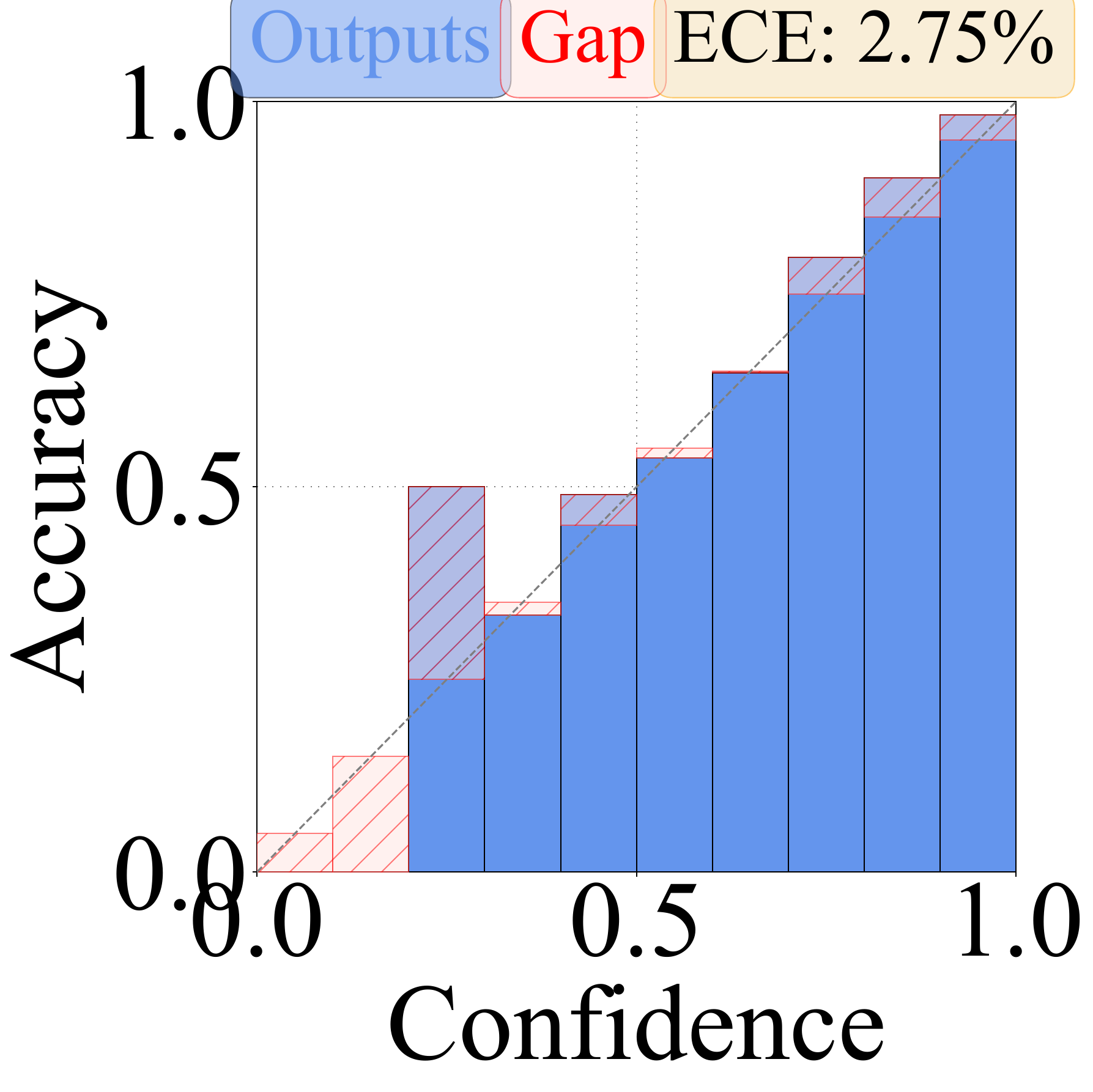}
    \caption{{Reliability plot for the proposed approach (\textbf{first row}) and baselines (LoRA, \textbf{second row}; PromptFL, \textbf{third row}) on ISIC2019 dataset. The clients and global are represented from the left column to the right column.}}
    \label{fig:ISIC2019_calibration}
\end{figure*}

\mypar{Impact of model compression.} We explored the impact of performance losses due to float 16 conversion during model compression on ISIC2019 dataset. As shown in Figure \ref{fig:T_Values}, the AVG is 80.48\% without compression, slightly higher than with compression. However, with compression, the model size is reduced from 2.01 (FAM itself) to 1.36MB. These results suggest that compressing the weights of the model does not considerably impact its performance.

\mypar{Sensitivity analysis.} We performed experiments using different $\lambda$ values on the ISIC2019 dataset to analyze its sensitivity. Figure \ref{fig:ISIC2019_ablation_attacks} shows the test ACC with various $\lambda$ values ($\lambda \in [0.01,0.1]$). As illustrated, $\lambda$ equal to 0.04 leads to the highest overall ACC (AVG=80.4) compared to other settings.

\mypar{Robustness.} We validate the robustness to gradient based adversarial attacks. Specifically, we consider the fast gradient sign method (FGSM) \cite{goodfellow2014explaining} and projected gradient descent (PGD) \cite{madry2018towards} for experiments on the ISIC2019 dataset. The magnitude of adversarial attacks is set to 0.1 for FGSM and PGD. As illustrated in Figure \ref{fig:ISIC2019_ablation_attacks}, PEFT approaches such as FedCLIP and LoRA exhibit a considerable performance decrease (e.g. $\sim 10\%$ AVG under FGSM attack), while the proposed method shows a higher AVG of 33.85\%, indicating its robustness to adversarial attacks compared to PEFT approaches. Similar situation can be found with PGD attack (e.g., the proposed method provides $\sim25\%$ higher ACC on $C_1$ compared to LoRA).

\mypar{Backbones.} We changed the network backbones with deeper/different architectures, such as ViT-L/14, ViT-H/14, ConvNext\_Large provided in OpenCLIP \cite{cherti2023reproducible,ilharco_gabriel_2021_5143773} and MambaOut-Base \cite{yu2025mambaout} to validate its adaptability. Figure \ref{fig:ISIC2019_ablation_attacks} shows the test ACC with these backbones. The proposed approach exhibits remarkable adaptability with deeper networks such as ViT-L/14 (81.83\% AVG) and MambaOut-Base (80.2\% AVG).

\mypar{Calibration.} Figure \ref{fig:ISIC2019_calibration} shows the reliability diagrams \cite{vaicenavicius2019evaluating} on the ISIC2019 dataset for the proposed method and baselines. As illustrated, the proposed approach demonstrates a lower ECE value on local clients (e.g., 1.93\% on $C_1$, 2.34\% on $C_3$ compared to LoRA (2.32\% on $C_1$ and 4.43\% on $C_3$) and PromptFL (3.25\% on $C_1$ and 2.52\% on $C_3$). In addition, our method achieves comparable ECE on global site (2.72\%) compared to LoRA (2.20\%) and PromptFL (2.75\%). These results suggest that the proposed framework provides a reasonable balance between classification and calibration, allowing reliable predictions.

\section{Conclusion}

In this study, we explored the potential of VLMs for medical imaging in FL and proposed a novel CLIP-based FL framework. We introduced a masked FAM as the communication module while freezing the CLIP encoders to reduce computational and communication overhead, while using masked MLPs to adapt the local client with class-wise KL. Finally, ensemble predictions are obtained to improve local performance. Experimental results in skin-, brain- and prostate-related classification tasks demonstrate the remarkable performance of our approach compared to SOTA methods.

\section{Acknowledgments}
This research was funded by the National Natural Science Foundation of China grant number 82260360, the Innovation Project of GUET Graduate Education 2025YCXS244 and the Guangxi Science and Technology Base and Talent Project (2022AC18004, 2022AC21040).

\bibliography{aaai2026}

\end{document}